\documentclass{article}

\usepackage{PRIMEarxiv}

\usepackage[utf8]{inputenc} 
\usepackage[T1]{fontenc}    
\usepackage{hyperref}       
\usepackage{url}            
\usepackage{booktabs}       
\usepackage{amsfonts}       
\usepackage{nicefrac}       
\usepackage{microtype}      
\usepackage{xcolor}         
\usepackage{graphicx}
\usepackage{subcaption}
\usepackage{enumitem}
\usepackage{amsmath}
\usepackage[numbers]{natbib}
\usepackage{tcolorbox}
\tcbuselibrary{breakable, skins}
\usepackage{xcolor}     

\usepackage{array}
\usepackage{wrapfig}
\usepackage{listings}
\usepackage{longtable}
\usepackage{float}
\usepackage{placeins}
\usepackage{fancyvrb}

\title{Masked Self-Distillation: Internalizing the Chain-of-Thought in Language Models}

\author{%
    Durgesh Kalwar\thanks{Equal contribution - Joint first Author}\\
	SCAI, Arizona State University\\
	\texttt{dkalwar@asu.edu} \\
	\And
    Vardhan Palod$^{*}$\\
	SCAI, Arizona State University\\
	\texttt{vpalod@asu.edu} \\
	\And
    Jaya Adithya Pavuluri \\
    SCAI, Arizona State University\\
	\texttt{jpavulu2@asu.edu} \\
	\And
	Subbarao Kambhampati \\
	SCAI, Arizona State University \\
	\texttt{rao@asu.edu} \\
}

\begin{document}

\maketitle

\begin{abstract}
Large Reasoning Models (LRMs) produce long, explicit chains of intermediate steps before generating a final answer at inference time. These intermediate traces dominate latency, memory usage, and serving cost, even though the final answer correctness is not causally related to the trace correctness and the trace length is not a reliable indicator of the problem complexity. This raises an obvious question: can the computation expressed in these intermediate tokens be internalized into the parameters of a language model, enabling it to produce answers with much shorter intermediate traces?
We propose \textbf{masked self-distillation}, a knowledge-distillation based post-training framework in which copies of the same model are instantiated as teacher and student, and the student model is trained to internalize all or part of the intermediate trace, thus becoming more efficient at inference. We vary the fraction of the intermediate trace the student is trained to internalize, interpolating between full internalization (the student emits only the solution) and no internalization (the student emits the full trace before the answer). We conduct controlled experiments on two reasoning domains: math and graph coloring. We use the masked self-distillation framework to post-train Qwen3-4B and Qwen3-8B models. Our results demonstrate that this method can be used to improve task performance while increasing inference efficiency across various domains and model sizes. We systematically analyze whether the improved efficiency gains in the post-trained models generalize to out-of-distribution (OOD) problems. We find that the masked self-distillation models generalize well for in-domain OOD problems, and the masked self-distillation training does not induce catastrophic forgetting in the student model on out-of-domain problems. Furthermore, our ablation study shows that supervised fine-tuning can train models to produce shorter traces, but at the cost of generalization, highlighting the importance of on-policy training in masked self-distillation. Lastly, we show that this framework can also be used effectively in a dual-model setting where a larger reasoning model teacher supervises a smaller non-reasoning student model, resulting in improved task performance and inference efficiency.
\end{abstract}

\section{Introduction}
Large Reasoning Models (LRMs), post-trained with RL, achieve strong performance on math, code, and planning benchmarks by generating long chains of intermediate tokens before producing a final answer. While these traces improve task performance, they also significantly increase inference cost. LRMs spend most of their generation budget on intermediate tokens rather than the answer, and inference latency, KV-cache footprint, and energy consumption all grow roughly linearly with trace length. At deployment, this is the main cost of serving them.

A broad line of work aims to reduce this cost for LRMs by adding trace length-control objectives during RL post-training [L1 \cite{aggarwal2025l1}, CoT-Valve \cite{ma2025cot}, O1-Pruner \cite{luo2025o1}, Kimi-1.5-style length-penalty rewards \cite{team2025kimi}, GFPO \cite{shrivastava2025sample}, \cite{arora2502training}]. These methods reward shorter traces, however, the trace length compression is dependent on the base model’s output distribution to generate a shorter trace during training. While the fact that performance increases with intermediate traces is well-known, the reasons for it are less clear. Recent work has shown that there is no causal connection between trace correctness and final-answer correctness \citep{valmeekam2025beyond, bhambri2025interpretable, chenreasoning, arcuschin2025chain}, and the trace length is not connected to the computational complexity of the problem instance being solved \citep{illusion-of-thinking,palod2025performative}. These findings raise an obvious question: If the model's reasoning is not causally connected to its solution, is it necessary for the model to explicitly produce these tokens? Can the computation expressed in these intermediate tokens be internalized into the parameters of a language model?

Building on \cite{hinton2015distilling, shenfeld2026self, hubotter2026reinforcement},
we study whether we can use knowledge distillation to internalize the information carried by intermediate tokens into the model's parameters. We propose \textbf{masked self-distillation}, a knowledge-distillation framework in which copies of the same model are instantiated as teacher and student, where the teacher is conditioned on the prompt and part of the reasoning trace, and the student is conditioned only on the prompt. The training objective is to minimize the reverse-KL divergence \citep{kullback1951information} between the teacher and student's probability distributions on responses generated by the student. This methodology enables the student to receive dense token-level feedback and leverage the benefits of RL on-policy exploration. 
Since the teacher is conditioned on a prefix of the reasoning trace, the student is incentivized to act as though the computations in that prefix were already in its context, which in turn leads it to produce shorter traces. We vary the percentage of trace provided to the teacher, thus controlling the amount of information the student has to internalize. 


We instantiate this framework for training Qwen3-4B and Qwen3-8B \citep{yang2025qwen3} models across two reasoning domains: math and graph coloring. We first investigate whether we can use this framework to post-train models to internalize the information present in their reasoning tokens. In both domains and across different models, our results demonstrate that masked self-distillation improves task performance while reducing inference cost.  We extensively investigate whether the efficiency gains generalize to out-of-distribution (OOD) problems. We conduct experiments in two OOD settings: (1) in-domain but outside the training distribution, and (2) out-of-domain - evaluating a model trained on graph coloring problems on math problems and vice versa. We find that the masked self-distillation models generalize well for in-domain OOD problems, and the masked self-distillation training does not induce catastrophic forgetting in the student model on out-of-domain problems.
Furthermore, we perform an ablation study to examine the role of reverse-KL on-policy training vs. supervised fine-tuning (SFT) in token reduction. We find that while SFT improves in-distribution performance, the masked self-distillation variants consistently outperform the SFT variants on OOD test sets. This suggests that the reverse-KL on-policy training incentivizes the models to learn efficient reasoning behavior rather than just memorizing shorter traces.
Finally, we also instantiate this framework in a dual-model setting where a larger reasoning model teacher supervises a different smaller non-thinking student model. Our studies show that the framework can also be effectively leveraged for training smaller models to improve task performance with efficient intermediate token generation.
\begin{figure}[htbp]
    \centering
    \includegraphics[width=0.9\linewidth, height=9cm, keepaspectratio]{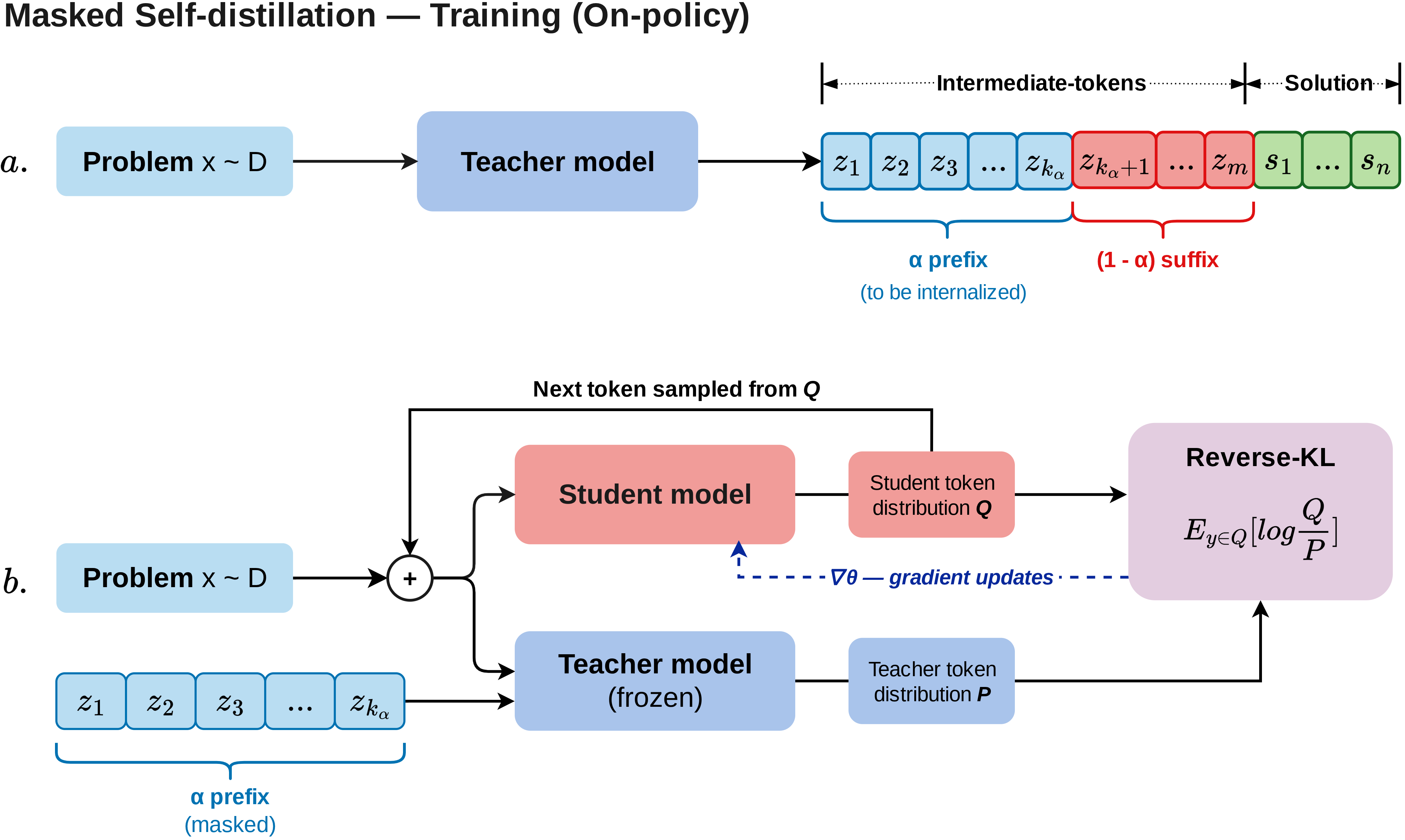}
    \caption{\textbf{Overview of the masked self-distillation framework.}
    \textit{a.} For each problem $x \sim \mathcal{D}$, we sample a response from the teacher model. From each response, we extract the intermediate tokens $z$ and the solution tokens $s$.
    \textit{b.} The teacher $\pi^{T}$ and the student $\pi^{S}_{\theta}$ are copies of the same model. The frozen teacher is conditioned on $x$ and the first $\alpha$ fraction of $z$, while the student is conditioned only on $x$, so it has to internalize this $\alpha$ fraction into its parameters. The student's distribution $Q$ is matched to the teacher's distribution $P$ with the on-policy reverse-KL loss $\mathbb{E}_{y \sim Q}[\log Q/P]$, computed on student-sampled responses; only the student is updated.}
\label{fig:masked-distillation}
\vspace{-6pt}
\end{figure}
\section{Related Work}
\textbf{Test-time compute and Inference cost reduction: }
\cite{DBLP:journals/corr/abs-2201-11903, kojima2023largelanguagemodelszeroshot} showed that LLMs perform significantly better on planning and reasoning tasks when they are prompted to produce intermediate steps before generating the final answer. Since then a new class of models called Large Reasoning Models (LRMs) have emerged that are trained to generate reasoning traces before giving the final answer \citep{jaech2024openai, comanici2025gemini, guo2025deepseek}. While there has been a significant performance increase in planning and reasoning domains, LRMs spend most of their generation budget on intermediate tokens. A broad line of token efficiency work has emerged that aims to reduce the intermediate token length at inference. During inference, there have been various prompting methods proposed to incentivize efficient reasoning behavior in an LRM \citep{han-etal-2025-token, zhao-etal-2026-thinking, wang-etal-2025-wait}. While they have shown some success in reducing inference cost, prompting methods depend heavily on the steerability of the LRM, and their efficacy is highly task-dependent. To induce efficient reasoning behavior in the model, there have been RL post-training methods that aim to reduce the length of the traces \citep{aggarwal2025l1, team2025kimi, shrivastava2025sample, arora2502training, luo2025o1, ma2025cot}. The general theme of these works is that during RL training, they reward responses with shorter traces and correct solutions sampled from the base model. These works have shown promising reductions in trace lengths. However, the compression is dependent on the base model's output distribution to generate a shorter trace during training for RL to reward them. In contrast, during the masked self-distillation training, a portion of the reasoning trace is masked from the student, and the student is trained to internalize the information in the fraction of the reasoning trace. Thus, in our method, there is a supervision signal that incentivizes the production of shorter traces at inference.

\textbf{Knowledge Distillation in LLMs: }
Knowledge distillation is widely studied to transfer information from a larger, more capable teacher model to a smaller student model \citep{hinton2015distilling, romero2015experimental, kim-rush-2016-sequence, sanh2019distilbert}. \cite{snell2022learning, gu2024minillm, agarwal2024policy} showed that this technique can be applied to LLM training as well. \cite{snell2022learning} introduced the concept of context distillation in a setting similar to supervised learning: the idea of sampling from a model provided with extra context and training the same model to mimic these predictions without that context. There have been works that used self-distillation to internalize additional context such as factual information or environmental feedback into the model's parameters \citep{scheurer2023training, dou2024re, zhou2026expo, kujanpaa2024efficient, shenfeld2026self, hubotter2026reinforcement}. While these works explore the internalization of additional context, they do not study it for LRMs and reducing the intermediate token length. 

\textbf{CoT compression and Latent reasoning: }Prior to the emergence of LRMs, some works investigated whether the Chain-of-Thought (CoT) tokens of an LLM can be internalized into the model's parameters \citep{snell2022learning, deng2023implicit, deng2024explicit}. These methods use an off-policy training objective, and while the in-distribution accuracy can usually be matched with increased inference efficiency, they do not investigate whether these gains are generalizable. Recently, there is an additional line of work that studies reasoning in latent space \citep{hao2024training, saunshi2025reasoning, su2025token}. These works suggest that the computations of the reasoning tokens can be compressed into the model parameters and thus, the models do not need to explicitly generate these tokens.



\section{Masked Self-Distillation}\label{sec:md_framework}

We present a knowledge-distillation framework, which we call \emph{masked self-distillation}, in which two copies of the same model are instantiated as a teacher and a student. The teacher copy of the model operates in think mode, while the student copy is initialized in no-think mode. The initial part (or all) of the trace is \emph{masked} from the student during the distillation i.e., the teacher sees it, but the student does not and the student model is trained to internalize the masked portion into its parameters. The higher the amount of trace the student internalizes, the shorter its outputs and the cheaper its inference, because that computation has moved into the student's parameters during distillation.

Figure~\ref{fig:masked-distillation} shows an overview of the \emph{masked self-distillation} framework. First, for each problem $x \in \mathcal{D}$, we sample a response from the teacher and take the segment between the \emph{<think>} and \emph{</think>} tags as the intermediate tokens $z = (z_1, \dots, z_m)$, where $m$ is the length of the reasoning trace. The simplest way to do masked distillation on the teacher response is to train the student with SFT on the suffix part of the intermediate tokens followed by the solution tokens. However, as SFT is an off-policy method, it is known to suffer from exposure bias \citep{pomerleau1991efficient, ross2010efficient} and does not generalize well \citep{chu2025sft, shenfeld2026self}. We therefore train the student using on-policy distillation. The student is only conditioned on $x$ and is trained to match the teacher's next-token distribution, conditioned on $x$ and a fractional prefix of $z$. We control what fraction of the trace is masked with a parameter $\alpha \in [0, 1]$. The teacher is conditioned on $x$ and the first $k_\alpha = \lfloor \alpha\, m \rfloor$ tokens of the trace, $z_{\le k_\alpha}$, while the student is conditioned on $x$ alone. Since the teacher is now effectively going to produce only the tokens after the $\alpha$ prefix, the student effectively has to internalize the $\alpha$ prefix of the trace. 

Let $y = (y_1, \dots, y_{L})$ be a response sampled from the student, where $L$ is the length of the response. The objective is to minimize the reverse-KL divergence between the student and teacher next-token distributions, computed on student-sampled responses:
\begin{equation}
\label{eq:md-loss}
\mathcal{L}^{\alpha}_{\mathrm{MSD}}(\theta)
  = \mathbb{E}_{\,x \sim \mathcal{D},\; y \sim \pi^{S}_{\theta}(\cdot \mid x)}
    \left[
      \sum_{t=1}^{L}
      D_{\mathrm{KL}}\!\left(
        \pi^{S}_{\theta}(\,\cdot \mid x, y_{<t})
        \,\big\|\,
        \pi^{T}(\,\cdot \mid x, z_{\le k_\alpha}, y_{<t})
      \right)
    \right],
\end{equation}
where, for next-token distributions $p$ and $q$ over the vocabulary $\mathcal{V}$,
\begin{equation}
\label{eq:kl}
D_{\mathrm{KL}}(q \parallel p) = \mathbb{E}_{v \sim q}\!\left[\log \frac{q(v)}{p(v)}\right]
= \sum_{v \in \mathcal{V}} q(v) \log \frac{q(v)}{p(v)}.
\end{equation}

Because $y$ is sampled from the student, the objective is on-policy: the student learns from the responses it actually generates, and it receives dense token-level feedback from the teacher at every position instead of a single sparse reward.

As $\alpha \in [0, 1]$, masked self-distillation has two extreme cases. In the first case ($\alpha = 1$ thus $k_\alpha = m$), the teacher is conditioned on question $x$ and the full reasoning trace, while the student is only conditioned on $x$. The student has to internalize the entire reasoning process and, at inference, emit only the solution tokens. 
We call this \emph{fully masked}, and the loss becomes
\begin{equation}
\label{eq:full-md-loss}
\mathcal{L}^{1}_{\mathrm{MSD}}(\theta)
  = \mathbb{E}_{x,\, y}
    \left[
      \sum_{t} D_{\mathrm{KL}}\!\left(
        \pi^{S}_{\theta}(\,\cdot \mid x, y_{<t})
        \,\big\|\,
        \pi^{T}(\,\cdot \mid x, z, y_{<t})
      \right)
    \right].
\end{equation}
In the second case ($\alpha = 0$, $k_\alpha = 0$), the teacher and the student are both conditioned only on $x$. 
We call this \emph{non-masked}, and the loss becomes
\begin{equation}
\label{eq:nonmasked-loss}
\mathcal{L}^{0}_{\mathrm{MSD}}(\theta)
  = \mathbb{E}_{x,\, y}
    \left[
      \sum_{t} D_{\mathrm{KL}}\!\left(
        \pi^{S}_{\theta}(\,\cdot \mid x, y_{<t})
        \,\big\|\,
        \pi^{T}(\,\cdot \mid x, y_{<t})
      \right)
    \right].
\end{equation}

\section{Experimental Setup}\label{sec:setup}
Our broad aim is to study how the masked self-distillation training affects inference cost and task performance. We conduct experiments in the following domains - 
\begin{enumerate}
    \item \textbf{Math -} We used the popular \textit{GSM8K} as our in-distribution dataset. This dataset, introduced by \citep{cobbe2021gsm8k}, consists of grade school math problems, and is designed to evaluate the reasoning capabilities of large language models. It contains 8.5K problems, each paired with a question and an answer. The dataset is divided into 7.5K training problems and 1K test problems. To test the out-of-distribution (OOD) generalization of models post-trained on GSM8K problems, we have in-domain OOD and out-of-domain test sets. We use AIME-25 \citep{aime25} and MATH-500 \citep{lightman2024let} as in-domain OOD test sets as these datasets share the same arithmetic substrate but contain problems of higher difficulty. For the out-of-domain test, we use the graph coloring $\mathrm{GC}$-$\mathrm{ER}_{\chi_{3-4}}$ test set (described below).
    
    \item \textbf{Graph Coloring -} 
    Graph coloring is a canonical NP-complete reasoning problem that is related to both constraint satisfiability as well as practical problems like scheduling and allocation. In this domain, the model is given a graph, specified by vertices and edges, and the number of possible colors and the task is to return a configuration where no two connected vertices have the same color. For a graph $G$, the minimum number of colors required to properly color its vertices is the chromatic number, denoted $\chi(G)$. We construct a custom dataset using the \texttt{networkx} Python library \citep{SciPyProceedings_11}, controlling problem difficulty by computing $\chi(G)$ for each sampled graph and retaining only those with the desired chromatic number.
    Our training set consists of $10{,}000$ problems on Erd\H{o}s--R\'enyi graphs \citep{erdds1959random} with $\chi(G) \in \{3,4\}$. The in-distribution test set consists of $500$ held-out problems. We use the notation $\mathrm{GC}$-$\mathrm{ER}_{\chi_{3-4}}$ to denote this set. For evaluating generalization, we have in-domain OOD and out-of-domain test sets. For in-domain generalization, we have test sets with harder problems {$\mathrm{GC}$-$\mathrm{ER}_{\chi_6}$} and {$\mathrm{GC}$-$\mathrm{BA}_{\chi_6}$} generated using the Erd\H{o}s--R\'enyi and Barab\'asi--Albert algorithms \citep{barabasi1999emergence} respectively (refer to appendix \ref{appx:dataset_detail} for more details). For the out-of-domain test, we use the \textit{GSM8K} test set.

\end{enumerate}


\textbf{Training details: } Our implementation builds on the \texttt{verl} library \citep{sheng2024hybridflow}. We train Qwen3-4B and Qwen3-8B models \citep{yang2025qwen3} and perform $\alpha$ masked self-distillation post-training in math and graph coloring domains. First, we generate responses from the teacher (base model in think mode) on the training data and then keep only the problems that the teacher solved correctly, i.e., rejection sampling (Figure~\ref{fig:masked-distillation}a). We perform the masked self-distillation training (Figure~\ref{fig:masked-distillation}b) on the student (base model in no-think mode) on the dataset created after rejection sampling. We vary $\alpha \in \{0.0, 0.3, 0.5, 0.7, 1.0\}$ to control how much of the trace the student internalizes and to study how this affects its accuracy and trace length. We provide additional hyperparameter details in the appendix section \ref{Appx:hyp}.
\section{Results and Discussion}\label{sec:results}


\subsection{Effect of Internalization on Task performance and inference efficiency}\label{subsec:internalization}
In this section, we study whether models post-trained using the masked self-distillation framework internalize the information present in the teacher’s reasoning tokens and achieve better task performance with improved inference efficiency. 

\begin{figure}[htbp]
    \centering
    \includegraphics[width=0.9\textwidth]{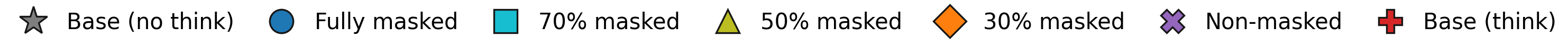}
    
    \begin{subfigure}{0.33\textwidth}
        \centering
        \includegraphics[width=\textwidth]{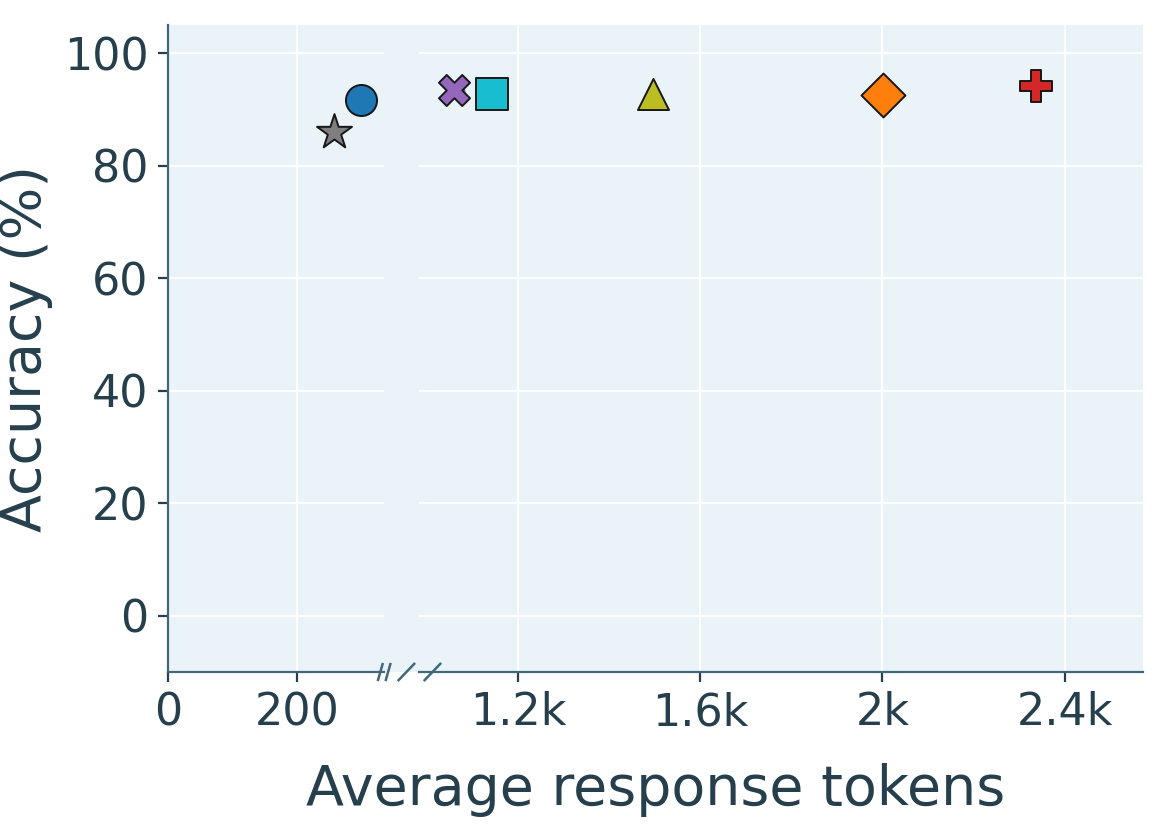}
        \caption{GSM8K}
        \label{fig:pareto-gsm8k-4b}
    \end{subfigure}\hfill
    \begin{subfigure}{0.33\textwidth}
        \centering
        \includegraphics[width=\textwidth]{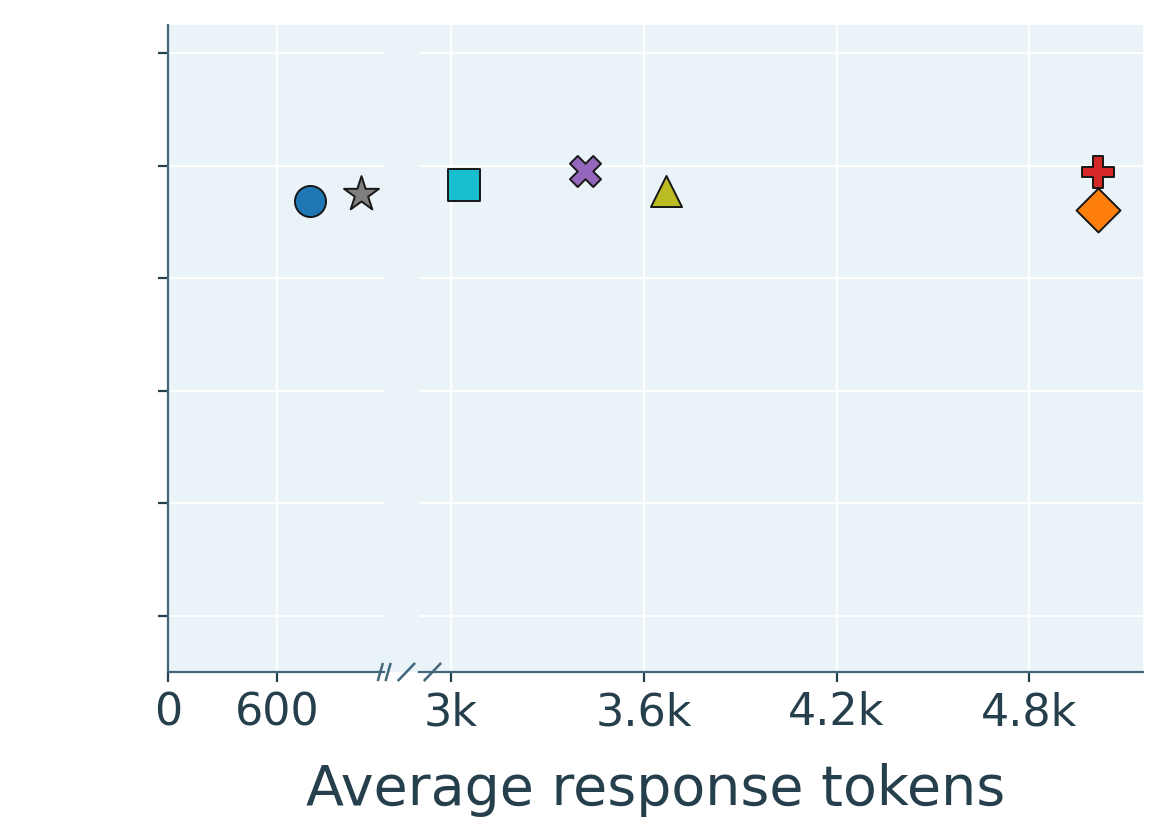}
        \caption{MATH-500}
        \label{fig:pareto-math-4b}
    \end{subfigure}\hfill
    \begin{subfigure}{0.33\textwidth}
        \centering
        \includegraphics[width=\textwidth]{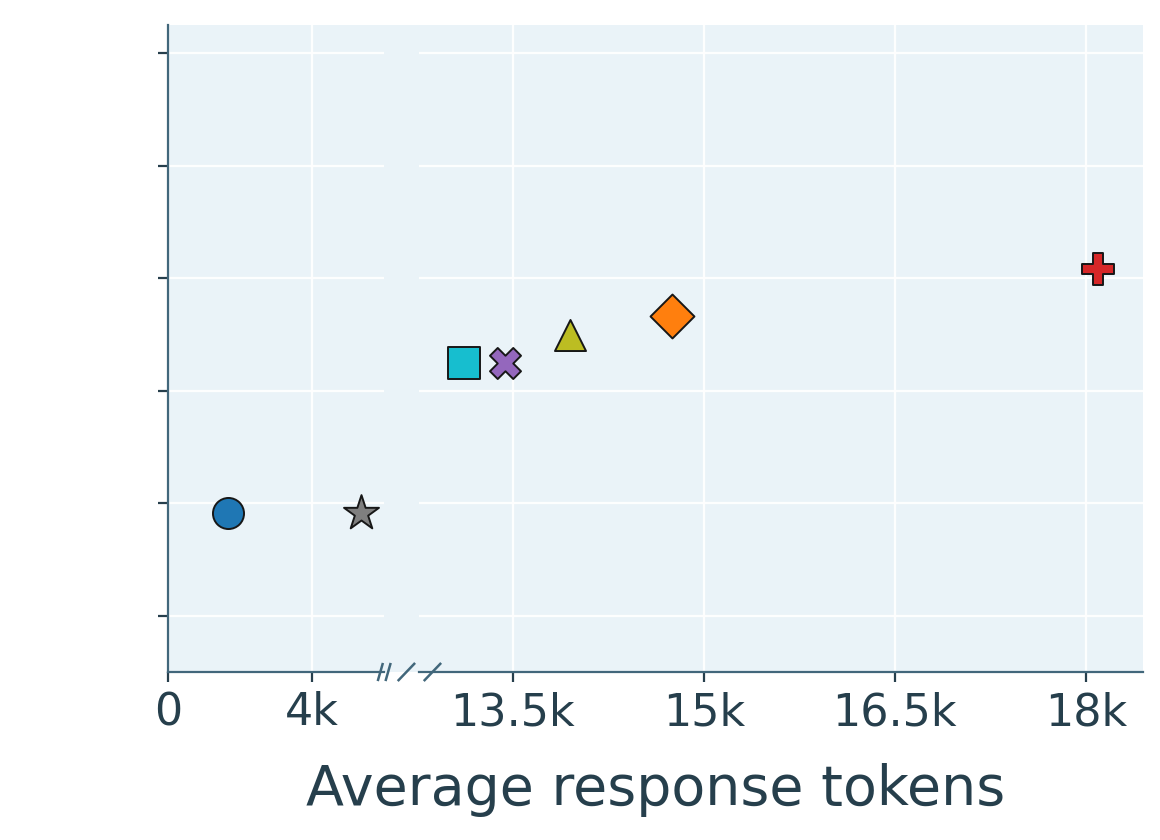}
        \caption{AIME25}
        \label{fig:pareto-aime-4b}
    \end{subfigure}
    
    \begin{subfigure}{0.33\textwidth}
        \centering
        \includegraphics[width=\textwidth]{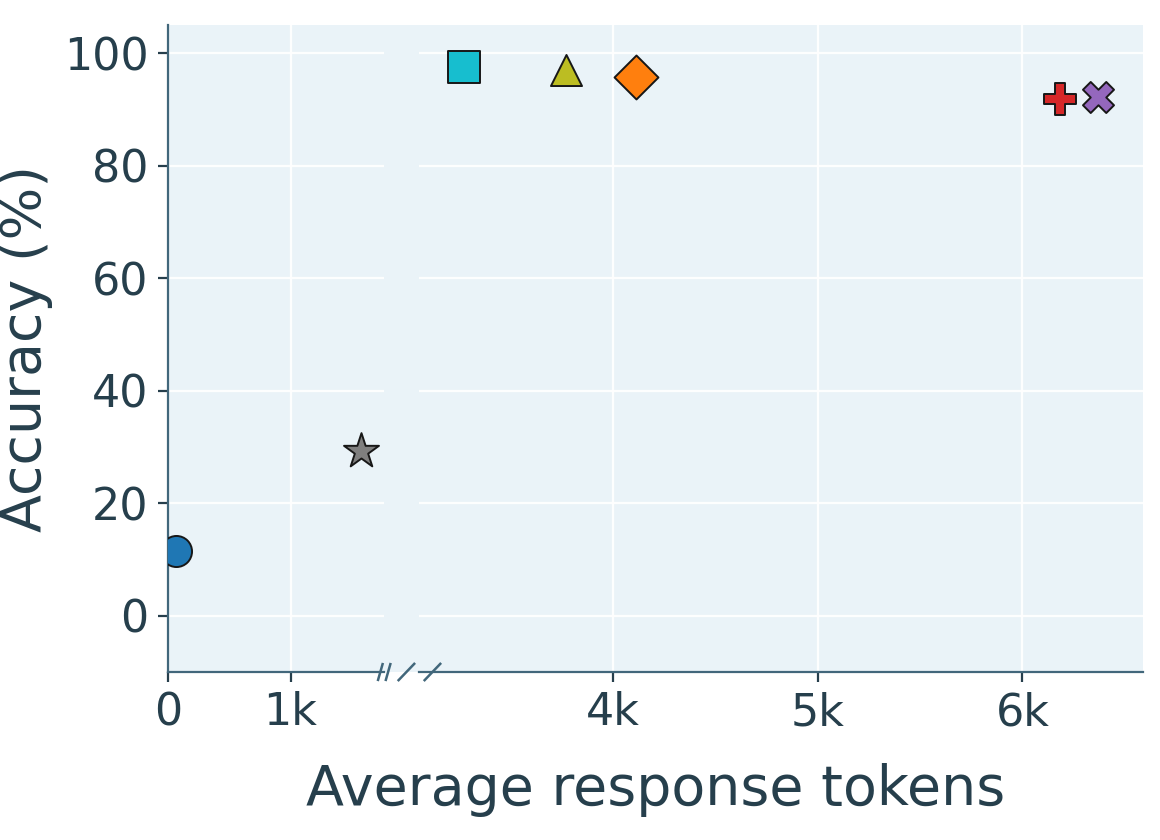}
        \caption{$\mathrm{GC}$-$\mathrm{ER}_{\chi_{3-4}}$}
        \label{fig:pareto-gc-er-id-4b}
    \end{subfigure}\hfill
    \begin{subfigure}{0.33\textwidth}
        \centering
        \includegraphics[width=\textwidth]{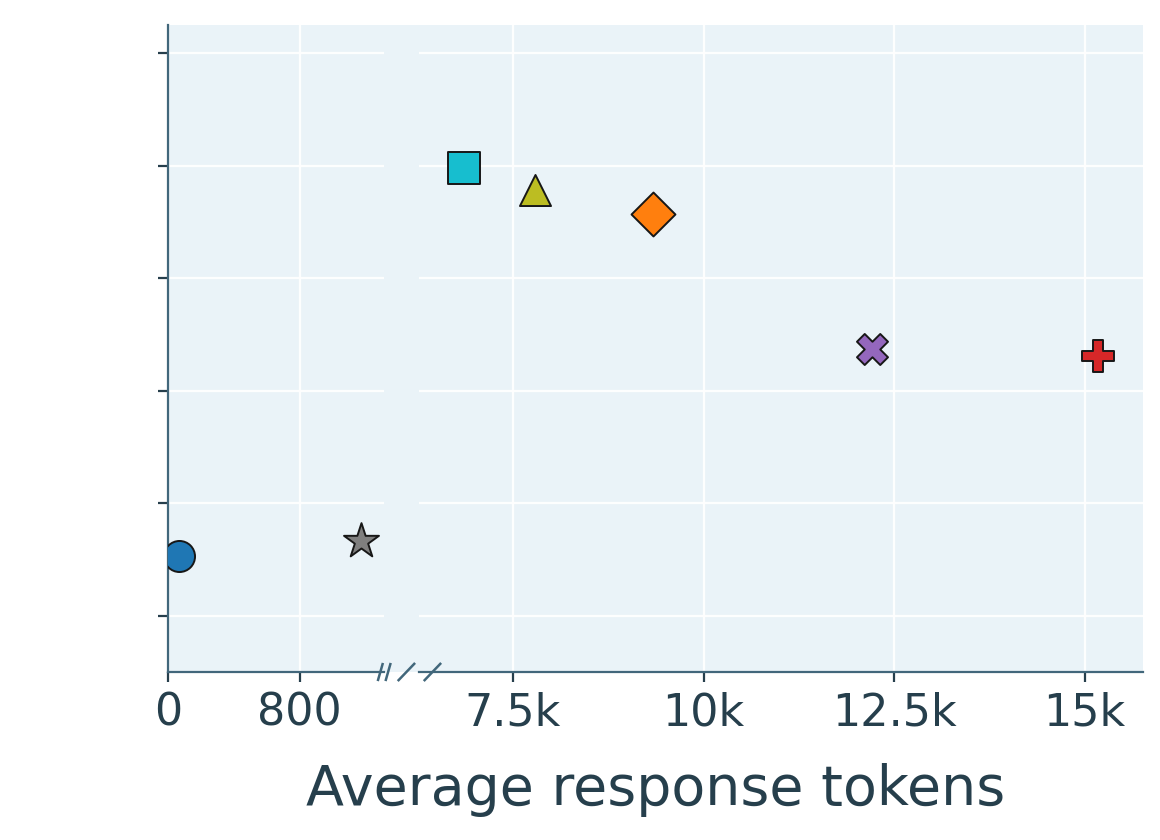}
        \caption{$\mathrm{GC}$-$\mathrm{ER}_{\chi_6}$}
        \label{fig:pareto-gc-er-ood-4b}
    \end{subfigure}\hfill
    \begin{subfigure}{0.33\textwidth}
        \centering
        \includegraphics[width=\textwidth]{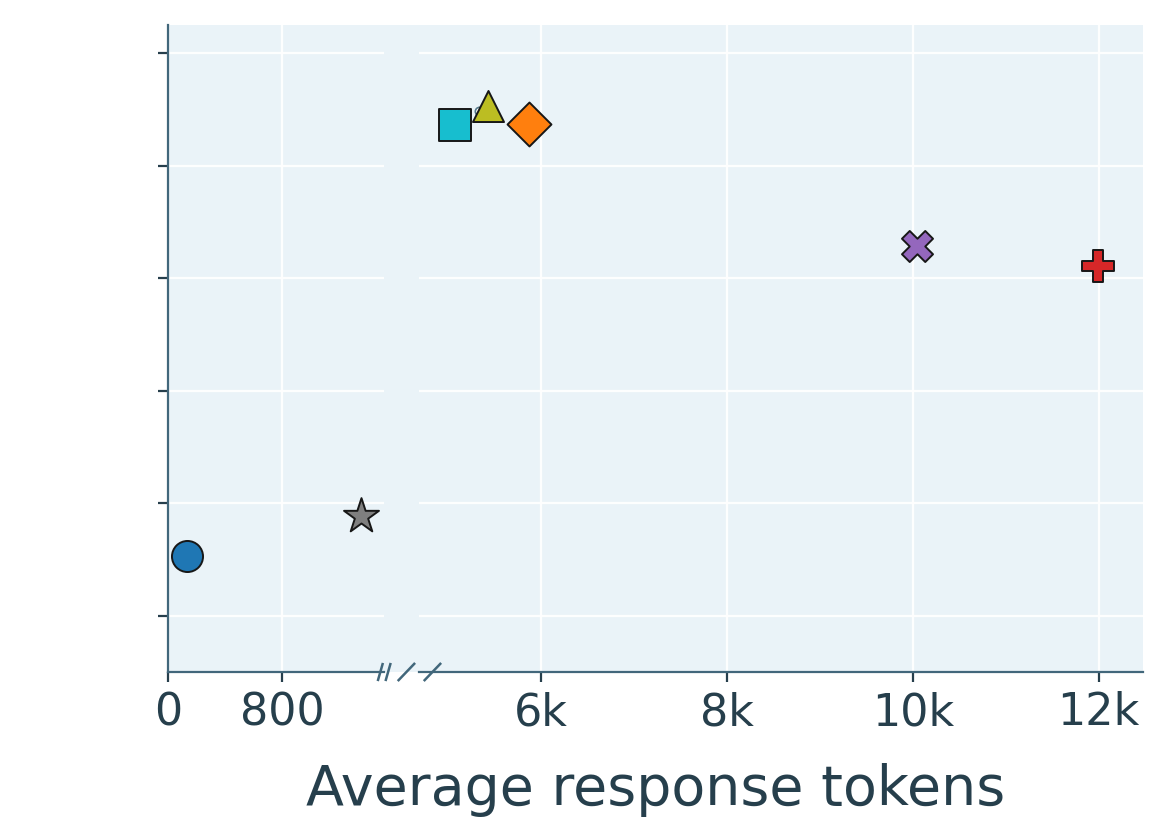}
        \caption{$\mathrm{GC}$-$\mathrm{BA}_{\chi_6}$}
        \label{fig:pareto-gc-ba-ood-4b}
    \end{subfigure}
    \caption{\textit{Qwen3-4B Pareto plot}: Accuracy vs.\ mean response length comparison on math (top row) and graph-coloring (bottom row) domain test-datasets for $\alpha$-masked variants, teacher (base think) and student (base no-think). Note: Accuracy and mean response are averaged over 3 random seeds.}
    \label{fig:pareto-self-distill-4b}
\end{figure}

\begin{figure}[htbp]
    \centering
    \includegraphics[width=0.9\textwidth]{figures/legends/regular_masking_legend_horizontal.png}
    \begin{subfigure}{0.33\textwidth}
        \centering
        \includegraphics[width=\textwidth]{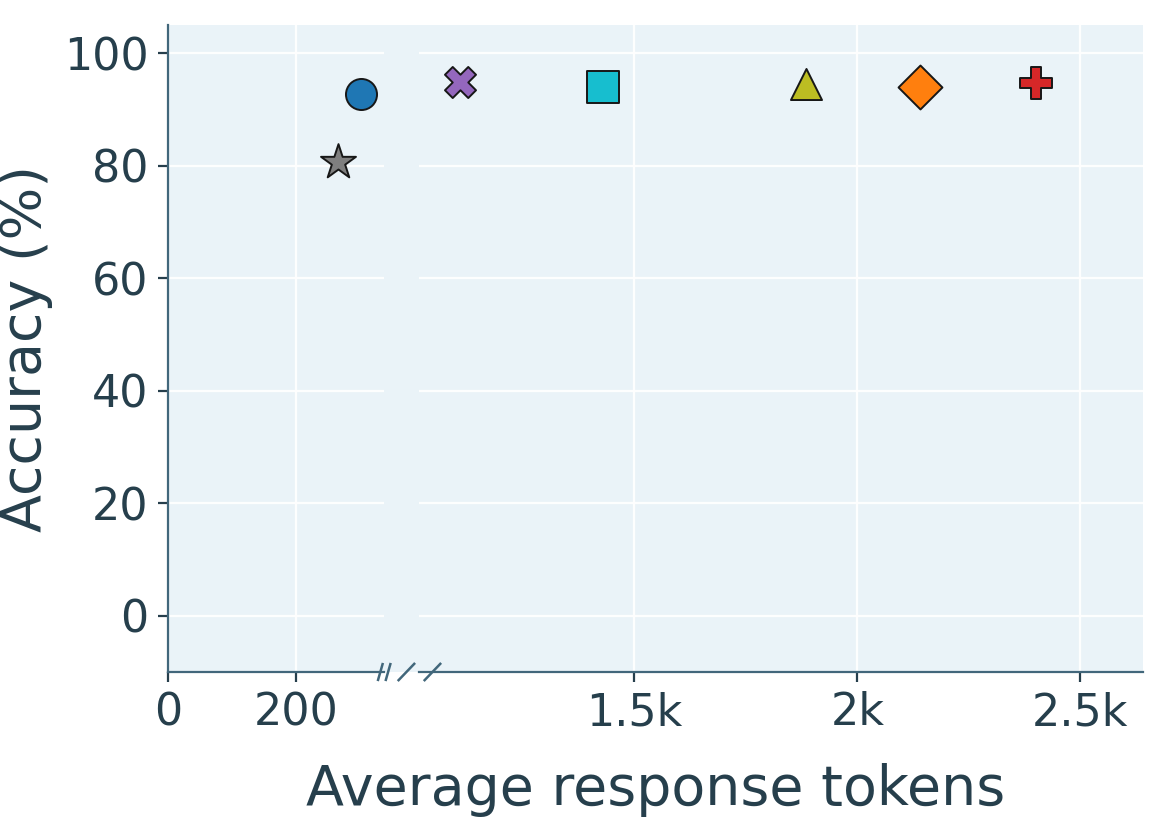}
        \caption{GSM8K}
        \label{fig:pareto-gsm8k-8b}
    \end{subfigure}\hfill
    \begin{subfigure}{0.33\textwidth}
        \centering
        \includegraphics[width=\textwidth]{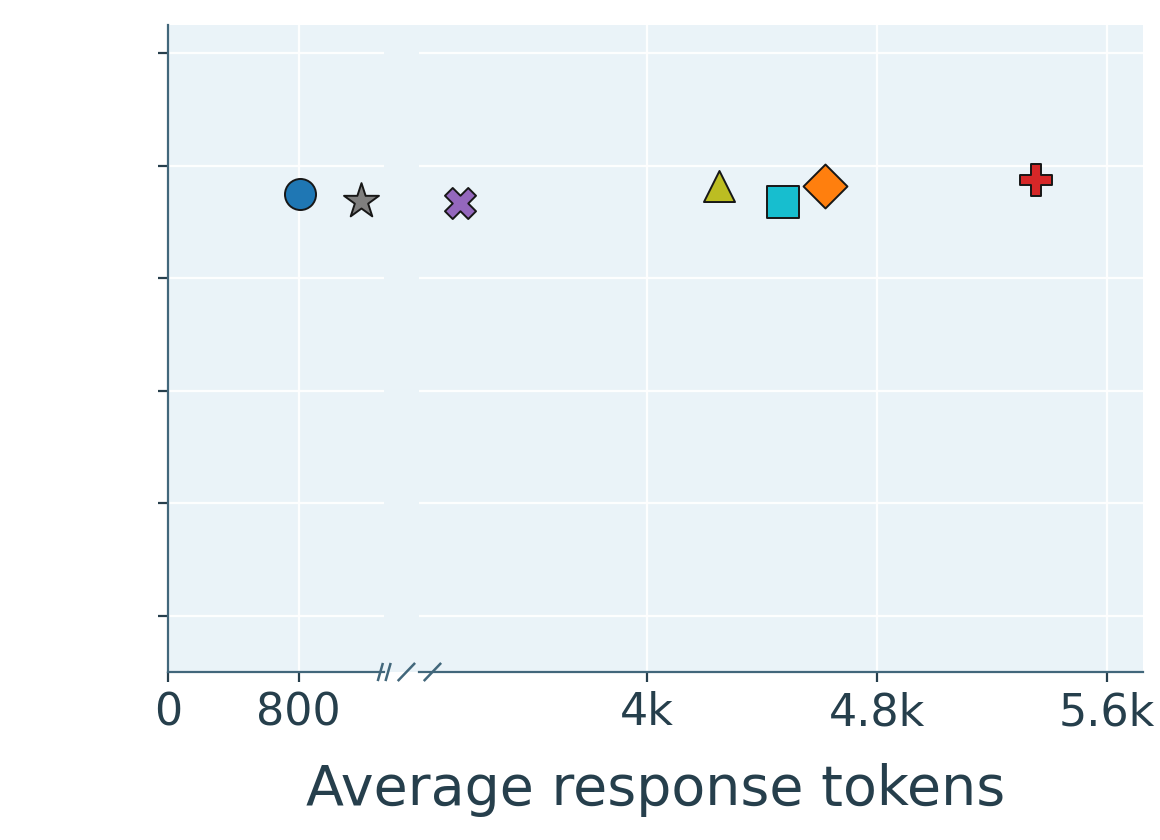}
        \caption{MATH-500}
        \label{fig:pareto-math-8b}
    \end{subfigure}\hfill
    \begin{subfigure}{0.33\textwidth}
        \centering
        \includegraphics[width=\textwidth]{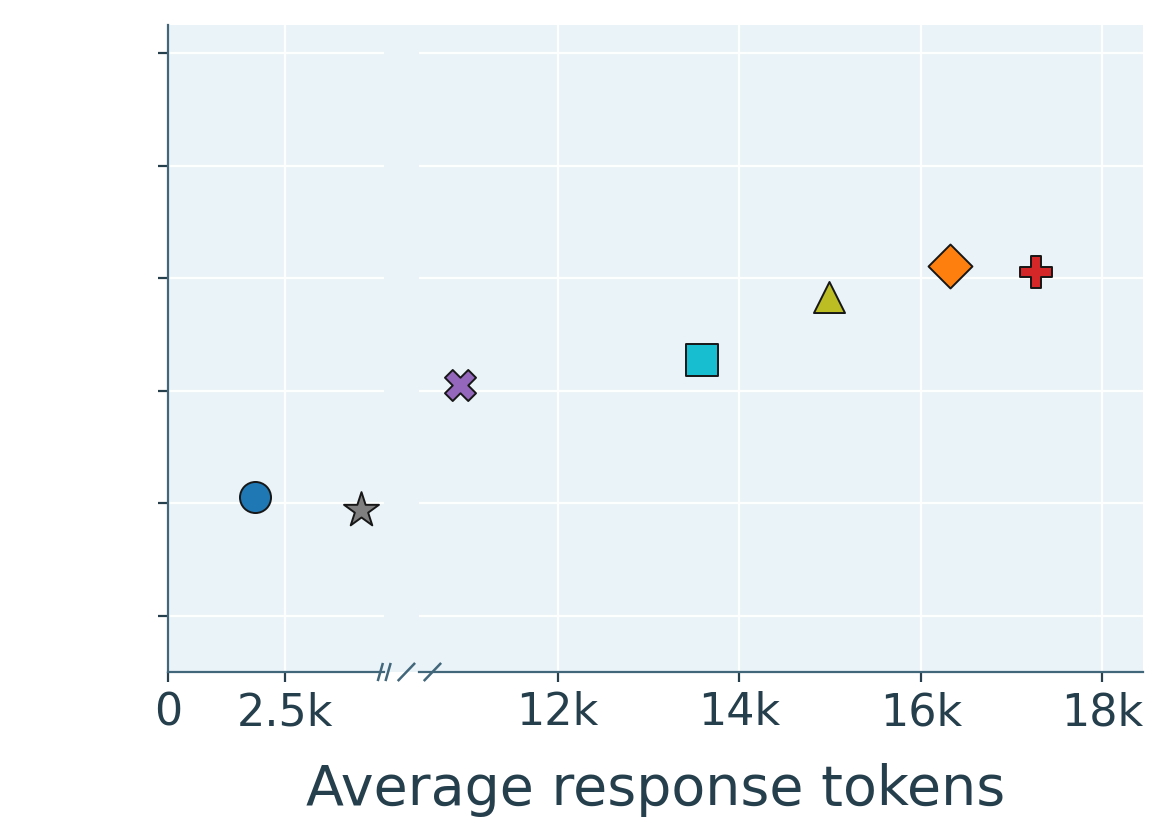}
        \caption{AIME25}
        \label{fig:pareto-aime-8b}
    \end{subfigure}
    
    \begin{subfigure}{0.33\textwidth}
        \centering
        \includegraphics[width=\textwidth]{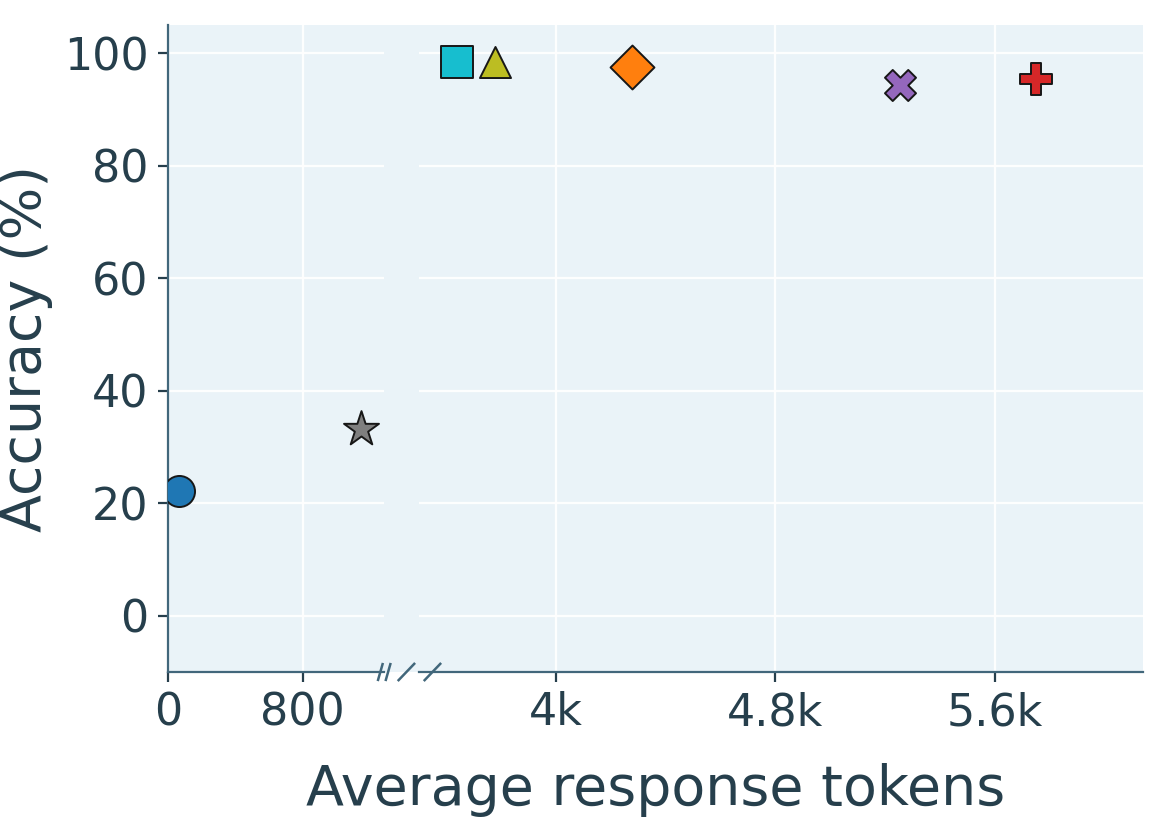}
        \caption{$\mathrm{GC}$-$\mathrm{ER}_{\chi_{3-4}}$}
        \label{fig:pareto-gc-er-id-8b}
    \end{subfigure}\hfill
    \begin{subfigure}{0.33\textwidth}
        \centering
        \includegraphics[width=\textwidth]{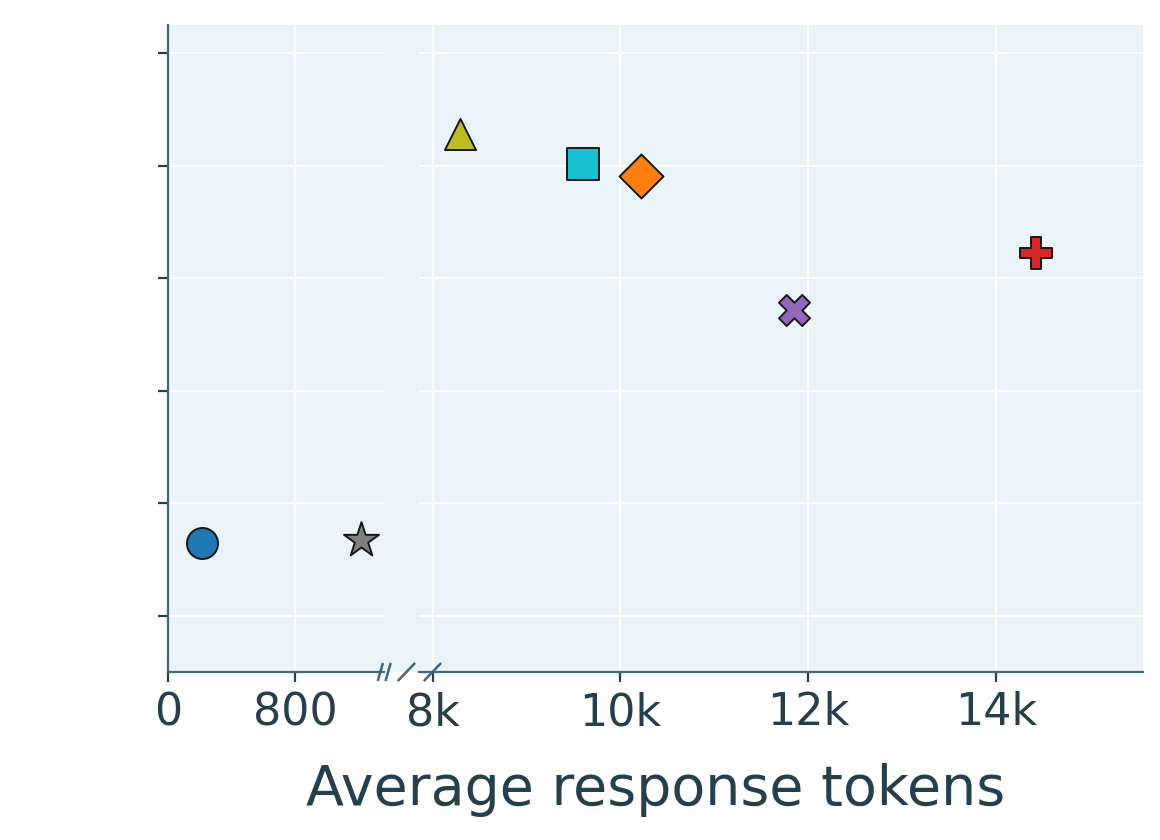}
        \caption{$\mathrm{GC}$-$\mathrm{ER}_{\chi_{6}}$}
        \label{fig:pareto-gc-er-ood-8b}
    \end{subfigure}\hfill
    \begin{subfigure}{0.33\textwidth}
        \centering
        \includegraphics[width=\textwidth]{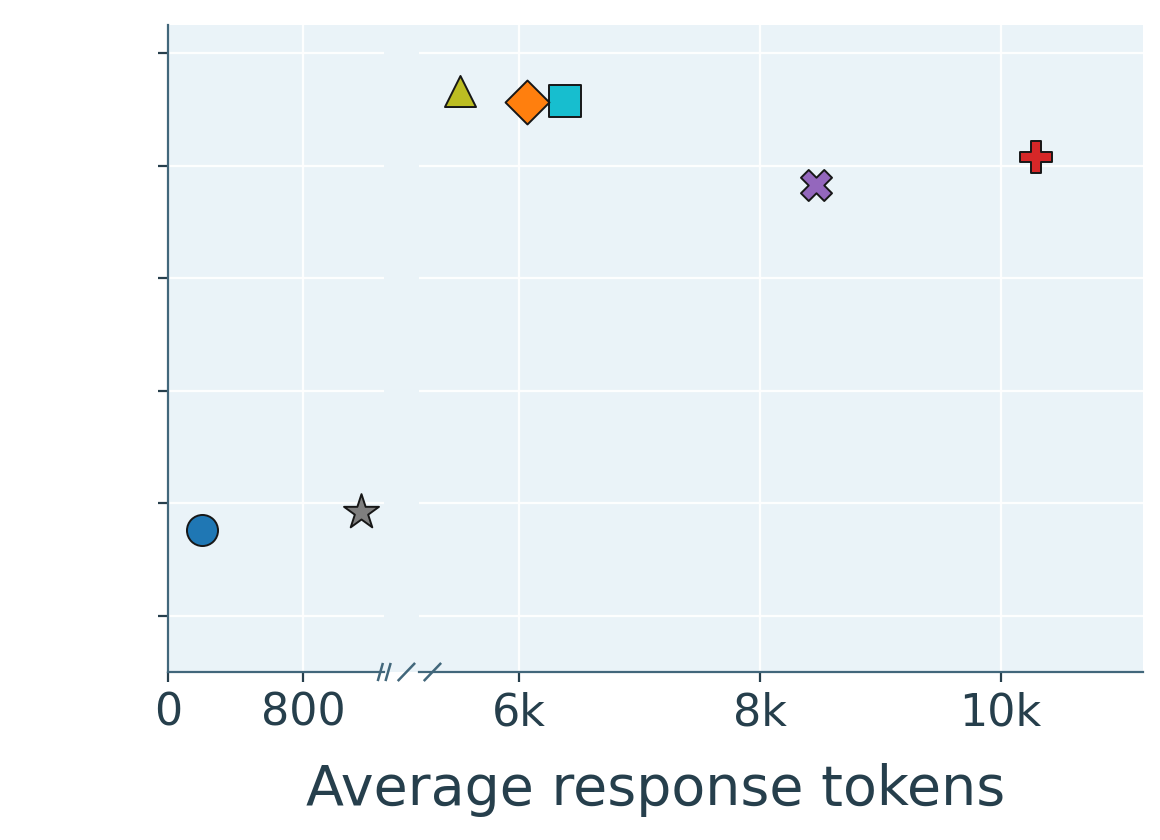}
        \caption{$\mathrm{GC}$-$\mathrm{BA}_{\chi_{6}}$}
        \label{fig:pareto-gc-ba-ood-8b}
    \end{subfigure}
    \caption{\textit{Qwen3-8B Pareto plot}: Accuracy vs.\ mean response length comparison on math (top-row) and graph-coloring (bottom-row) domain test-datasets for $\alpha$-masked variants, teacher (base think) and student (base no-think). Note: Accuracy and mean response are averaged over 3 random seeds.}
    \label{fig:pareto-self-distill-8b}
\end{figure}

Figure~\ref{fig:pareto-self-distill-4b} shows accuracy vs.\ average response length for \emph{masked self-distillation} across different values of $\alpha$, together with the base student and teacher models, averaged over 3 random seeds \footnote{We report the confidence intervals in appendix \ref{app:4b_multi_seed} and \ref{app:8b_multi_seed}.}, on the in-distribution and out-of-distribution datasets for the math and graph coloring domains for the Qwen3-4B model. Figure~\ref{fig:pareto-self-distill-8b} shows the same comparison for the Qwen3-8B model. 

As shown in Figure~\ref{fig:pareto-gsm8k-4b}, the student (base no-think) accuracy is 86\% on the GSM8K test set while producing $\sim$260 tokens on average, while the teacher's (base think) accuracy is 94\% and it produces $\sim$2.3k tokens on average. After we train the student models on the GSM8K training set with masked self-distillation for different values of $\alpha$, they produce far fewer tokens than the teacher while keeping similar performance. The fully masked variant ($\alpha = 1$) gives the best trade-off: it reaches $\sim$92\% accuracy with about $8\times$ fewer tokens than the teacher. 


On the graph coloring in-distribution test set (Figure~\ref{fig:pareto-gc-er-id-4b}), the student model accuracy is 29\%, and it generates $\sim$1.5k tokens on average, while the teacher reaches 92\% with $\sim$6k tokens. After training with masked self-distillation, the student produces shorter outputs, and the variants with $\alpha \in \{0.3, 0.5, 0.7\}$ exceed the teacher by 4–5 points, while producing 1.5-2x fewer tokens. Unlike on GSM8K, the fully masked training is not effective here. The fully masked model produces only $\sim$70 tokens, but its accuracy drops to 12\%, even less than the base student model. This suggests that fully internalizing the reasoning process is not feasible for the graph-coloring domain, and the student still requires a scaffold of intermediate tokens to perform well.

We observe two patterns in the average response length across different values of $\alpha$. (1) In the math domain, the trace generated by the $\alpha$-masked variants becomes shorter as $\alpha$ increases, with one exception: the non-masked variant ($\alpha = 0$) produces about as many tokens as the $\alpha = 0.7$ variant. (2) In the graph coloring domain, the variants with $\alpha \in \{0.3, 0.5, 0.7\}$ form a cluster and produce nearly the same number of tokens, within a few hundred tokens of each other, while the non-masked variant produces the most tokens among the trained variants. We attribute this behavior to the mode-seeking nature of the training objective \citep{gu2024minillm, agarwal2024policy}. Since we train the student with on-policy distillation, minimizing the reverse-KL divergence between the student's and the teacher's token distributions, the objective strongly penalizes the student for placing probability mass where the teacher's probability mass is small or zero. The student therefore concentrates on the high-density regions (modes) of the teacher and ignores its low-probability tails. The teacher can have several such modes, and which one the student converges to depends on the trajectories it samples during training. In masked self-distillation, the student is trained to behave as though a fraction of the reasoning trace were already in its context, and it tends to generate the remaining part of the trace, but this may not correspond to the exact trace the teacher would have generated. If the student happens to generate shorter or longer traces during training, and the teacher assigns high probability to them, the student can converge to a teacher mode that favors traces of that length. We believe this is what happens in both domains: for $\alpha = 0$ in the math domain, the student converges to a mode that produces traces almost as short as those of the $\alpha = 0.7$ variant, while for $\alpha = 0$ in the graph coloring domain, it converges to a mode that produces even more tokens than the teacher on the in-distribution test set.

The Qwen3-8B model shows the same pattern in both domains (Figures~\ref{fig:pareto-gsm8k-8b} and~\ref{fig:pareto-gc-er-id-8b}). On GSM8K, the fully masked variant reaches 93\% accuracy, close to the teacher's 95\%, with about $8\times$ fewer tokens. On graph coloring, the variants with $\alpha \in \{0.3, 0.5, 0.7\}$ match the teacher's accuracy of 95\% while producing about $1.4$--$1.6\times$ fewer tokens. 

While it is extremely difficult to do a systematic study of the content of the reasoning traces expressed in natural language, our qualitative analysis results suggest that the traces produced by the $\alpha$-masked variants seem disconnected, as if beginning
from an intermediate point in the reasoning (refer to appendix sections \ref{app:trace-analysis} and \ref{app:trace-eg} for more details). Taken together, these results suggest that the student models indeed learn to internalize the computations present in the $\alpha$ fraction of the teacher’s trace and thus become more efficient at inference.

\subsection{Analyzing the out-of-distribution effectiveness}\label{subsec:ood}
We want to evaluate whether the masked self-distillation training leads to memorization of shorter traces on in-distribution (ID) problems at the expense of generalization. We extensively study the out-of-distribution (OOD) performance in two OOD settings: (1) in-domain but outside the training distribution, and (2) out-of-domain.

\textbf{In-domain OOD:} On the math-domain OOD datasets, MATH-500 and AIME-25 (Figures~\ref{fig:pareto-math-4b} and~\ref{fig:pareto-aime-4b}), we observe the same trend as on GSM8K. On MATH-500, all $\alpha$-masked variants match the teacher's accuracy while producing far fewer tokens. On AIME-25, the $\alpha = 0.3$ variant reaches about 55\% accuracy, close to the teacher, while generating about $1.3\times$ fewer tokens. The remaining $\alpha$-masked variants are less accurate than the teacher but are more inference-efficient. 

The graph coloring OOD results are extremely promising (Figures~\ref{fig:pareto-gc-er-ood-4b} and~\ref{fig:pareto-gc-ba-ood-4b}). The teacher reaches only 45\% accuracy on $\mathrm{GC}$-$\mathrm{ER}_{\chi_{6}}$ and 62\% on $\mathrm{GC}$-$\mathrm{BA}_{\chi_{6}}$, while the best $\alpha$-masked variant ($\alpha = 0.7$) reaches 80\% on $\mathrm{GC}$-$\mathrm{ER}_{\chi_{6}}$ and 89\% on $\mathrm{GC}$-$\mathrm{BA}_{\chi_{6}}$. This is $\sim$30 percentage points above the teacher, with about half as many tokens. We attribute this to the on-policy nature of masked self-distillation: the student is trained explicitly on this domain, on trajectories generated by its own policy.

The in-domain OOD results for Qwen3-8B are consistent with those for Qwen3-4B (Figure~\ref{fig:pareto-math-8b}--\ref{fig:pareto-aime-8b} and Figure~\ref{fig:pareto-gc-er-ood-8b}--\ref{fig:pareto-gc-ba-ood-8b}) on both domains. Taken together, these results indicate that a student trained with masked self-distillation generalizes robustly on in-domain OOD problems, matching the teacher's performance in the math domain and surpassing it in the graph coloring domain, while being more inference-efficient. 

\textbf{Out-of-domain Generalization:} Since we train the student models on a single domain (math or graph coloring), a natural question is whether the resulting gains in both accuracy and inference cost carry over to other domains, or whether the models overspecialize to the training domain and lose accuracy elsewhere. To test this, we perform out-of-domain evaluation tests.

\begin{figure}[htbp]
    \centering

    \includegraphics[width=0.8\textwidth]{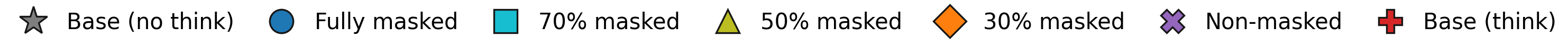}

    \vspace{0.5em}
    \begin{subfigure}[t]{0.45\textwidth}
        \centering
        \includegraphics[width=\textwidth]{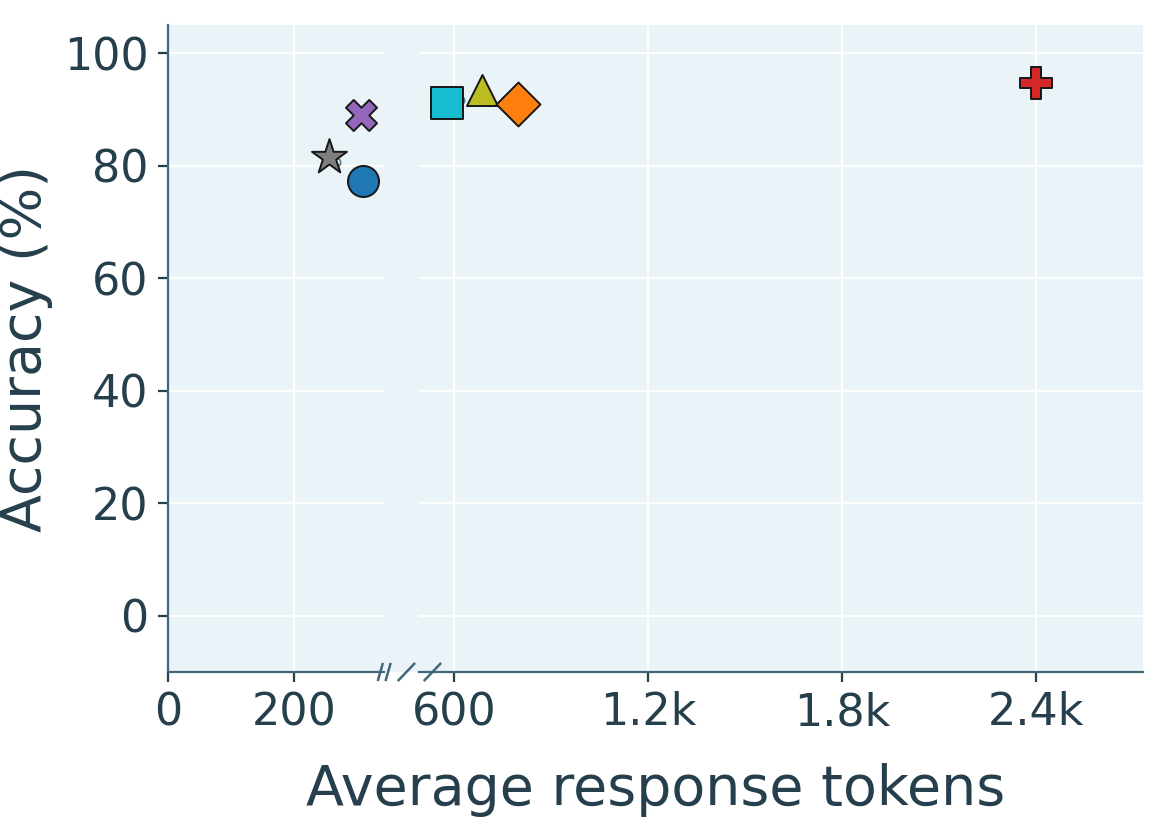}
    \end{subfigure}%
    \hspace{0.05\textwidth}%
    \begin{subfigure}[t]{0.45\textwidth}
        \centering
        \includegraphics[width=\textwidth]{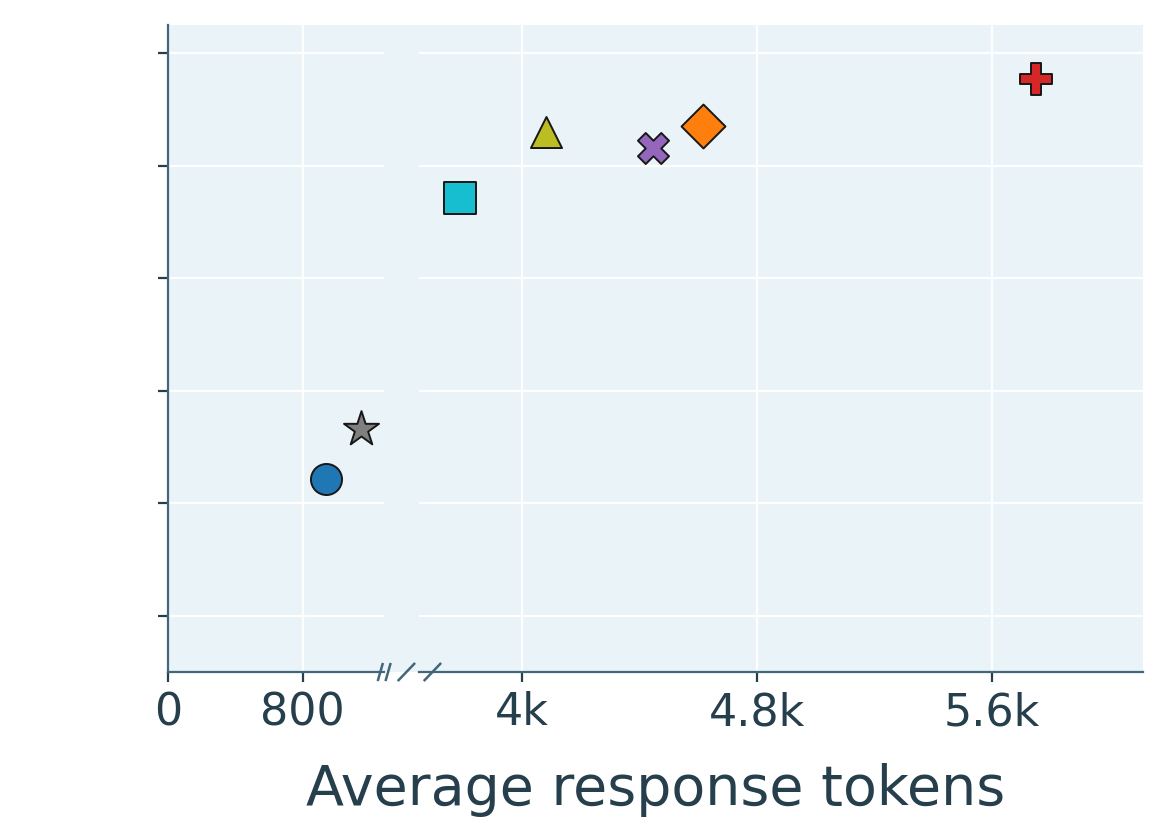}
    \end{subfigure}

    \caption{\textit{Out-of-domain Qwen3-8B Pareto plots:} (left) graph coloring models evaluated on the GSM8K test set and (right) GSM8K models evaluated on $\mathrm{GC}$-$\mathrm{ER}_{\chi_{3-4}}$ test set. }
    \label{fig:cross-domain-8b}
\end{figure}

Figure~\ref{fig:cross-domain-8b} shows out-of-domain evaluation analysis for $\alpha$-masked Qwen3-8B models, together with the base student and teacher models. The models trained on the graph coloring domain retain their inference cost savings even on the GSM8K test dataset. However, the models trained on the GSM8K dataset have slightly lower performance on $\mathrm{GC}$-$\mathrm{ER}_{\chi_{3-4}}$ test set compared to the base teacher model. On the $\mathrm{GC}$-$\mathrm{ER}_{\chi_{3-4}}$ test set, the teacher model reaches 95.5\% accuracy with $\sim$5.7k tokens, while the best masked self-distillation variant ($\alpha = 0.3$) reaches 87\% with $\sim$4.6k tokens, and $\alpha = 0.5$ reaches 85.5\% with $\sim$4.2k tokens. These variants remain far above the base student model (33\%), so the training still improves performance on this harder task, but the token cost reduction is minimal. These results show that while the inference cost gains are domain-dependent, the masked self-distillation training does not induce catastrophic forgetting in the student models. We report the out-of-domain analysis for $\alpha$-masked Qwen3-4B model in Appendix figure \ref{fig:cross-domain-4b}.



\subsection{Effect of the maximum token budget}

\begin{wrapfigure}{r}{0.42\textwidth}
    \centering\
    \includegraphics[width=0.35\textwidth]{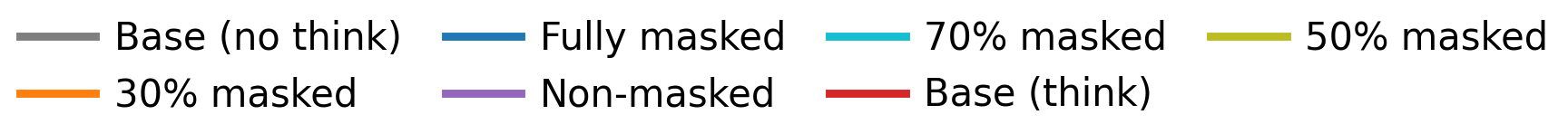}
    \includegraphics[width=0.35\textwidth]{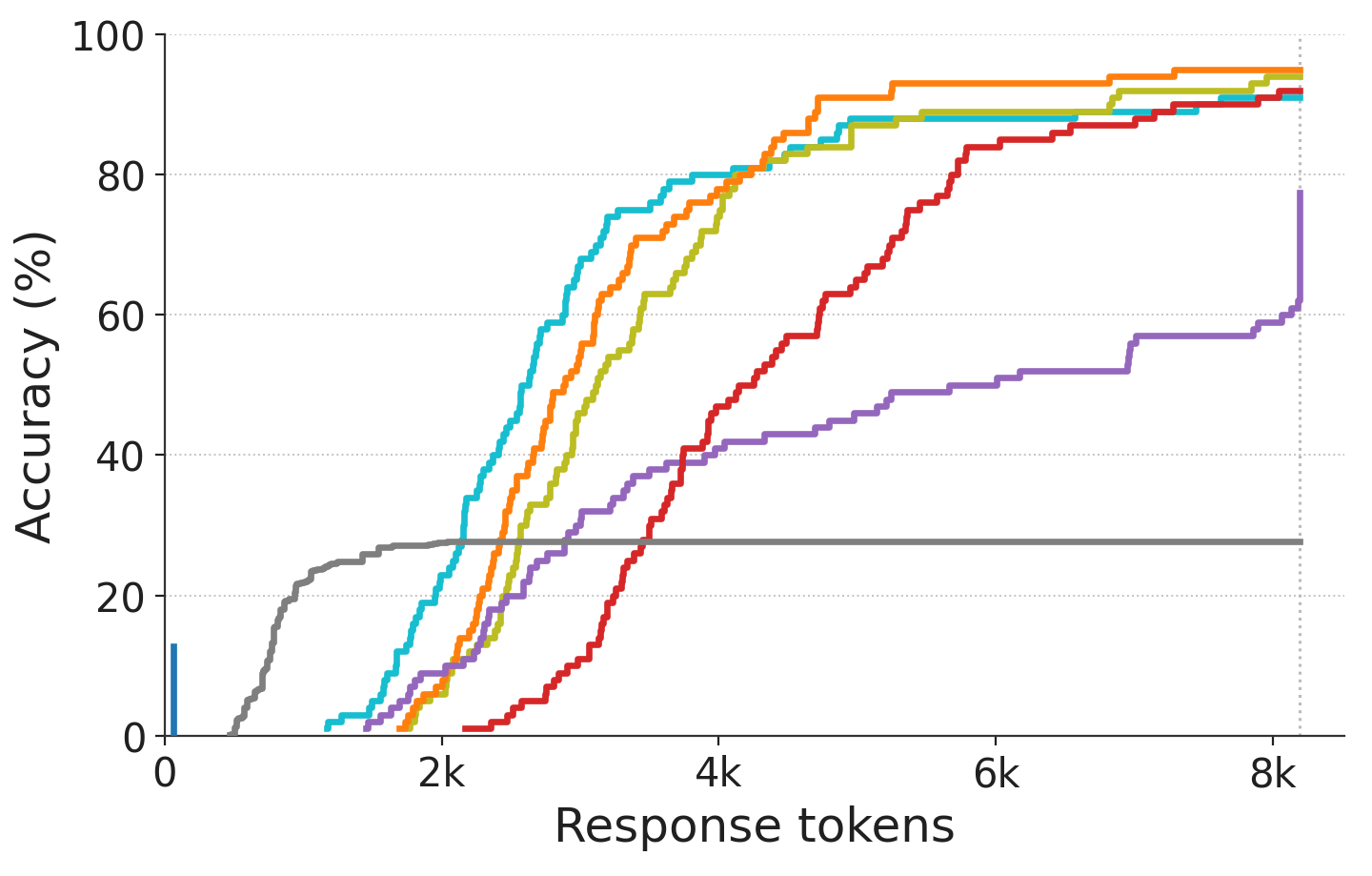}
    \caption{\textit{Qwen3-4B}: Accuracy curve over max-token budget.}
    \label{fig:learning-gc-er-id-4b-8k}
\end{wrapfigure}


The Pareto plots report a single operating point per baseline, namely its accuracy and mean token length \footnote{The mean response length does not provide a complete picture of the response-length distribution over the test set. To examine how masked distillation changes the response-length distribution and what distributions the $\alpha$-masked variants follow, we report violin plots in Appendix~\ref{subsec:length-dist}.} under a fixed cap. To see the full trade-off, we also plot the accuracy curve over the token budget: for each budget $b$ on the $x$-axis, the curve reports the accuracy the model would achieve if generation were stopped at $b$ tokens. A problem counts as solved only if the model produces a correct answer within $b$ tokens. Figure~\ref{fig:learning-gc-er-id-4b-8k} shows these curves for all baselines on the $\mathrm{GC}$-$\mathrm{ER}_{\chi_{3-4}}$ test set under an 8k cap for Qwen3-4B model. Masked self-distillation shifts the entire curve to the left. The curves for $\alpha \in \{0.3, 0.5, 0.7\}$ lie above and to the left of the teacher's over the whole range. This demonstrates that the $\alpha$-masked variants perform better than the base teacher at every fixed token budget. Notably, at a budget of 4k tokens, the $\alpha = 0.3$ variant already solves about 80\% of the problems, while the teacher solves fewer than 50\%. 

\subsection{Ablation Study: SFT vs on-policy distillation}
We compare the $\alpha$-masked variants trained with SFT and masked self-distillation to examine whether the effectiveness of our framework comes from the internalization of the teacher's reasoning traces together with the student's on-policy exploration for shorter reasoning traces, or whether similar gains can be achieved simply through over-training using SFT on a fraction of the teacher's reasoning trace and the final solution. We provide the details of the SFT training procedure in the appendix~\ref{Appx:sft_hp}.

\begin{table}[htbp]
\centering
\small
\begin{tabular}{lcccccccc}
\toprule
& \multicolumn{2}{c}{$\mathrm{GC}$-$\mathrm{ER}_{\chi_{3-4}}$} & \multicolumn{2}{c}{$\mathrm{GC}$-$\mathrm{ER}_{\chi_{6}}$} & \multicolumn{2}{c}{$\mathrm{GC}$-$\mathrm{BA}_{\chi_{6}}$} & \multicolumn{2}{c}{GSM8K}\\
\cmidrule(lr){2-3}\cmidrule(lr){4-5}\cmidrule(lr){6-7}\cmidrule(lr){8-9}
Masked & SFT & Reverse KL & SFT & Reverse KL & SFT & Reverse KL & SFT & Reverse KL \\
\midrule

100\%   & \textbf{88.7} & 11.5 & \textbf{11.4} & 10.5 & \textbf{26.2} & 10.6 & 0.0 & \textbf{87.0} \\
        & {\tiny 68} & {\tiny \underline{67}}
        & {\tiny 68} & {\tiny \underline{66}}
        & {\tiny \underline{68}} & {\tiny 115}
        & {\tiny 32k$^*$} & {\tiny \underline{299}} \\

70\%    & 94.2 & \textbf{97.5} & 52.7 & \textbf{79.5} & 72.1 & \textbf{88.3} & 68.5 & \textbf{87.7} \\
        & {\tiny \underline{2631}} & {\tiny 3274}
        & {\tiny \underline{5899}} & {\tiny 6906}
        & {\tiny \underline{4681}} & {\tiny 5185}
        & {\tiny 4866} & {\tiny \underline{549}} \\

50\%    & 95.1 & \textbf{97.0} & 60.2 & \textbf{75.5} & 82.3 & \textbf{89.9} & 77.0 & \textbf{90.8} \\
        & {\tiny \underline{3431}} & {\tiny 3772}
        & {\tiny \underline{7816}} & {\tiny 7986}
        & {\tiny 5631} & {\tiny \underline{5410}}
        & {\tiny 9084} & {\tiny \underline{1051}} \\

30\%    & 94.1 & \textbf{95.8} & 58.5 & \textbf{71.1} & 82.7 & \textbf{87.3} & 82.2 & \textbf{89.6} \\
        & {\tiny 4211} & {\tiny \underline{4112}}
        & {\tiny 10299} & {\tiny \underline{9384}}
        & {\tiny 6661} & {\tiny \underline{5942}}
        & {\tiny 5501} & {\tiny \underline{933}} \\

0\%     & \textbf{95.4} & 92.3 & \textbf{55.9} & 46.7 & \textbf{70.1} & 65.7 & 84.8 & \textbf{92.6} \\
        & {\tiny \underline{5433}} & {\tiny 6369}
        & {\tiny 13369} & {\tiny \underline{12439}}
        & {\tiny 10248} & {\tiny \underline{10186}}
        & {\tiny 4606} & {\tiny \underline{1670}} \\

\bottomrule
\end{tabular}
\caption{\textit{Qwen3-4B variants }: Accuracy and mean token length comparison between SFT vs. on-policy trained $\alpha$-masked variants. Each cell reports accuracy (\%) on the first line and mean token length on the second. Bold marks the highest accuracy; underline marks the shortest mean response. "*" denotes the model collapses and exhausts its token generation budget.}
\label{tab:sft_vs_kd}
\end{table}
\vspace{-3pt}

Table~\ref{tab:sft_vs_kd} reports the accuracy and mean response length of the SFT and masked self-distillation trained $\alpha$-masked variants. We post-train the Qwen3-4B model in no-think mode on variants of the graph-coloring training set where each variant has question, an $\alpha$ fraction of the trace, and the solution. The models are evaluated on the graph-coloring in-domain OOD test sets and on the out-of-domain GSM8K test set. From the table, we observe that for $\alpha \in \{0.3,0.5,0.7\}$, the variants trained with the on-policy reverse-KL objective generalize better to both the in-domain OOD and out-of-domain test sets than their SFT counterparts. For the fully masked variant ($\alpha=1$), SFT appears to overfit to the training distribution: although it performs well on the in-distribution test set, its performance drops substantially on the OOD test sets. Moreover, on the out-of-domain test set, the SFT-trained fully masked variant collapses, whereas the reverse-KL-trained fully masked variant largely maintains the performance of the base model. 
Furthermore, across all values of $\alpha$, the SFT-trained variants generate substantially longer responses on the cross-domain test set than their reverse-KL counterparts. This suggests that simply training on shortened teacher traces through SFT is not sufficient for the student to consistently learn shorter reasoning behavior. In contrast, reverse-KL training combines the internalization of the teacher's reasoning trace with on-policy trajectories generated by the student itself, leading to shorter responses while better preserving generalization. 

\subsection{Ablation Study: Dual Model Setting}


In our \emph{masked self-distillation} framework, we use copies of the same model as student and teacher, where the student is in no-think mode and teacher in think mode. But our framework can be instantiated in dual model setting as well, in which a larger reasoning model operates as teacher and supervises a smaller non-thinking student model. We evaluate the \emph{dual-model} setting on graph coloring, where teacher is Qwen3-8B (in think mode) and student is Qwen3-1.7B (in no-think mode).

\begin{figure}[!ht]
    \centering
    \includegraphics[width=0.9\textwidth]{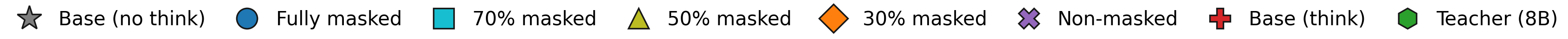}
    \begin{subfigure}{0.33\textwidth}
        \centering
        \includegraphics[width=\textwidth]{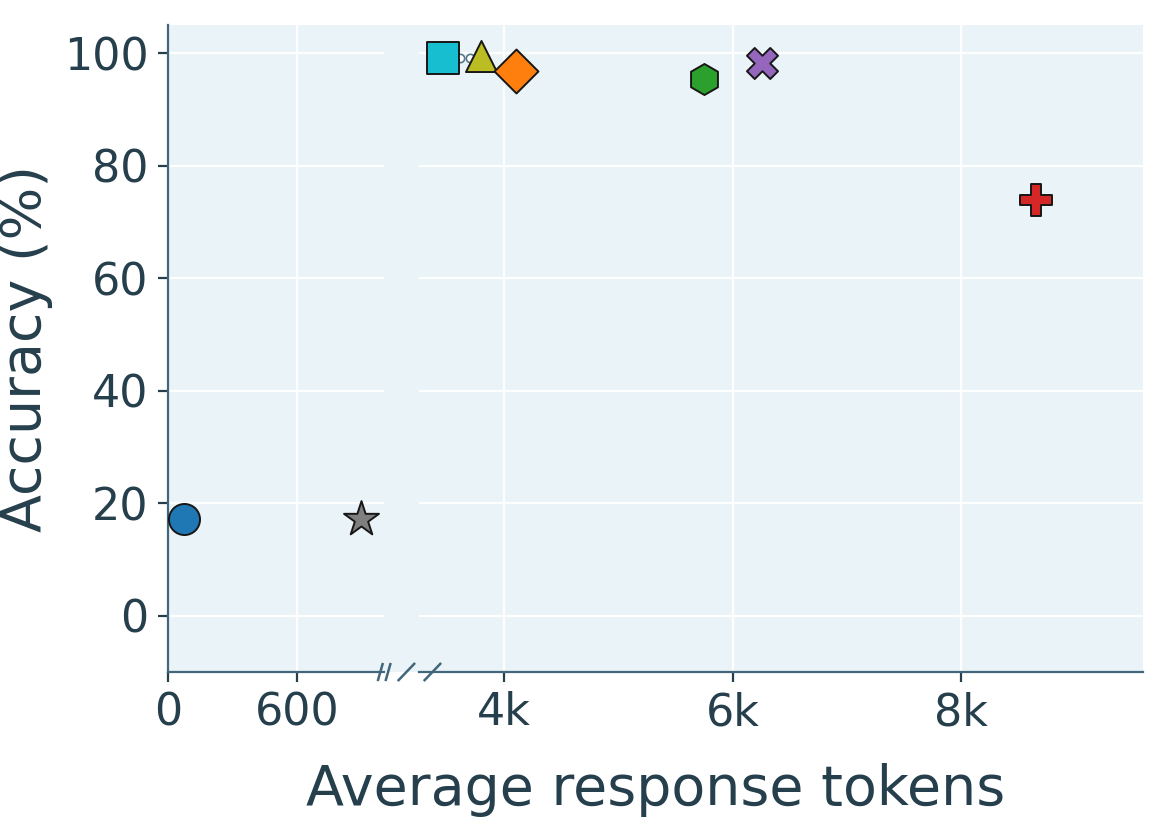}
        \caption{$\mathrm{GC}$-$\mathrm{ER}_{\chi_{3-4}}$}
        \label{fig:pareto-gc-er-id-dual}
    \end{subfigure}\hfill
    \begin{subfigure}{0.33\textwidth}
        \centering
        \includegraphics[width=\textwidth]{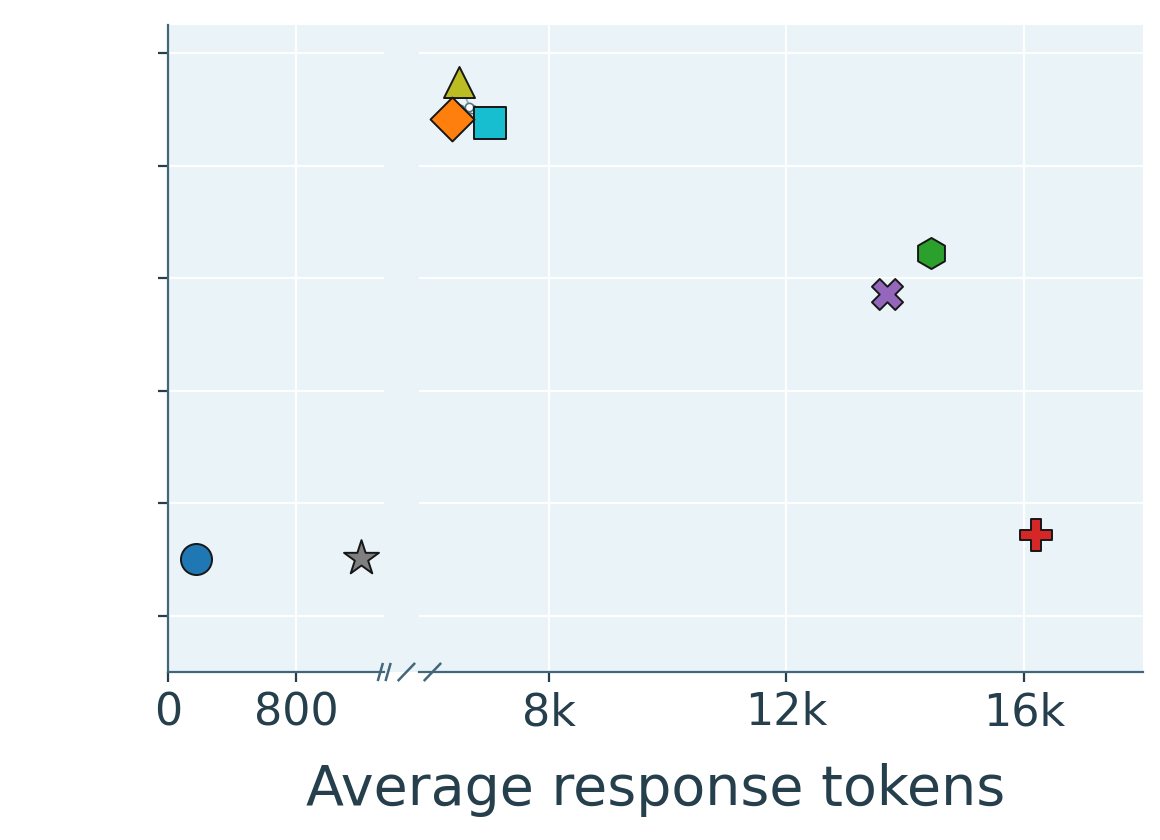}
        \caption{$\mathrm{GC}$-$\mathrm{ER}_{\chi_{6}}$}
        \label{fig:pareto-gc-er-ood-dual}
    \end{subfigure}\hfill
    \begin{subfigure}{0.33\textwidth}
        \centering
        \includegraphics[width=\textwidth]{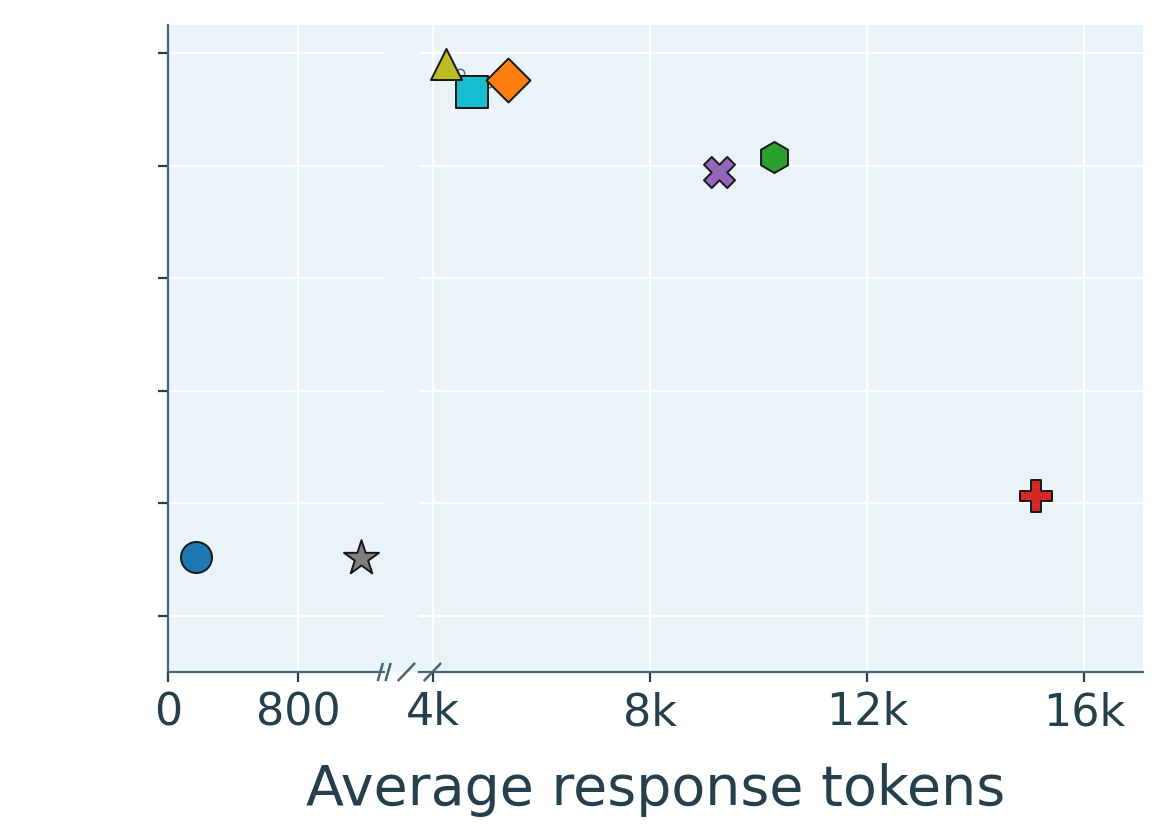}
        \caption{$\mathrm{GC}$-$\mathrm{BA}_{\chi_{6}}$}
        \label{fig:pareto-gc-ba-ood-dual}
    \end{subfigure}
    \caption{\textit{Dual-model setting Pareto plot}: accuracy vs.\ average response length comparison on graph-coloring domain test datasets for $\alpha$-masked variants, base think, student (base no-think) and teacher (8B think).}
    \label{fig:pareto-dual-model-gc}
\end{figure}

Figure \ref{fig:pareto-dual-model-gc} shows accuracy vs.\ mean token length for \emph{dual-model} setting across different values of $\alpha$, together with the base student model, teacher model, and the base model in reasoning mode on the in-distribution and out-of-distribution datasets. We observe a trend similar to that in the self-distillation setting. The $\alpha$-masked variants, $\alpha \in \{0.3,0.5,0.7\}$, internalize a fraction of the teacher's reasoning traces and generate substantially fewer tokens on average compared to the base reasoning model while achieving accuracies that are 30-40 percentage points higher on the in-distribution and both out-of-distribution test datasets. These results demonstrate that the \emph{masked distillation} framework can also be used to post-train smaller models to reason more efficiently.
\section{Conclusion}\label{sec:conclusion}

In this work, we study whether the computation expressed in the intermediate tokens can be internalized into the parameters of a language model. We proposed \textit{masked self-distillation}, a knowledge-distillation based post-training method that can be utilized to induce token-efficient reasoning without any external feedback or supervision.  Our results across math and graph coloring demonstrate that masked self-distillation consistently improves task performance while reducing intermediate tokens. We also conducted experiments in the countdown domain and find similar results (refer to Appendix \ref{app:countdown-results}). We extensively studied the generalization effectiveness of the masked self-distillation training. Our findings demonstrate that models post-trained using the framework generalize well on in-domain OOD problems and do not suffer from catastrophic forgetting on out-of-domain problems. We also compare with SFT trained models and show how the on-policy reverse-KL objective trains the models to generate shorter intermediate traces rather than just memorizing the shorter trace completions. Finally, we show that the masked distillation training can be effectively leveraged for training smaller models to improve task performance with efficient intermediate token generation.

\section*{Acknowledgments}
This research is supported in part by grants from ONR grant N00014-25-1-2301, ARO grant W911NF-26-1-A278, and gifts from Qualcomm and Apple. We also thank ASU AIR (Gaudi2) platform for the compute. 

\bibliography{references}
\bibliographystyle{plainnat}

\newpage
\appendix
\clearpage
\tableofcontents
\newpage
\section{Hyperparameter details}\label{Appx:hyp}
\subsection{Masked Self-Distillation}\label{Appx:msd_hp}
In our masked-self-distillation experiments, the student uses a maximum prompt length of $1{,}024$ tokens and a maximum response length of $8{,}192$ tokens. The teacher uses a longer context length of $10{,}240$ tokens to accommodate the fraction of intermediate tokens (ITs), collected in Phase~1 along with the question. Each training step uses a question batch size of $16$. The total number of training steps is 200. We observe that the models are trained to convergence. For the 4B models, we use a mini-batch size of $1$ and $8$ rollouts per question. For the 8B models, we use $5$ rollouts per question. The mini-batch size remains $1$ for GSM8K and is increased to $2$ for the 8B graph-coloring experiments. We optimize with AdamW at a constant learning rate of $1\times10^{-6}$, with no warmup and weight decay $0.01$. For all runs, we clip the actor gradient L2 norm at $1$. Rollouts at training time are sampled with the vLLM inference engine at temperature $1.0$, top-$p=1.0$, and top-$k=-1$. All experiments are run on a single node using between $2$ and $8$ Intel Gaudi2 HPUs \cite{10.1145/3785462.3815814}, depending on the model size and task.

\subsection{Dual-model setting}\label{Appx:dual_hp}
The dual-model experiments use Qwen3-1.7B as the student and a frozen Qwen3-8B teacher. We use a question batch size of $16$, a mini-batch size of $1$, and $5$ rollouts per question. The optimizer, training steps, sequence lengths, rollout sampling parameters, and gradient clipping follow the masked self-distillation setting above. All dual-model setting experiments are run on a single node with $4$ Intel Gaudi2 HPUs \cite{10.1145/3785462.3815814}.
\FloatBarrier

\subsection{SFT}\label{Appx:sft_hp}
We additionally study a supervised fine-tuning (SFT) variant of masked distillation, in which the student is trained on teacher collected responses (Figure~\ref{fig:masked-distillation}a) using the standard cross-entropy loss. For teacher generated responses, let $z$ be the intermediate trace and $s$ the solution tokens that follow it. The SFT target for a given $\alpha$ is the suffix of the trace followed by the solution, $y^{\alpha} = (z_{k_\alpha+1}, \dots, z_m, s)$, and the student is conditioned only on $x$:
\begin{equation}
\label{eq:md-sft-loss}
\mathcal{L}^{\alpha}_{\mathrm{SFT}}(\theta)
  = -\,\mathbb{E}_{\,x \sim \mathcal{D},\; (z, s) \sim \pi^{T}(\cdot \mid x)}
    \left[
      \sum_{t=1}^{|y^{\alpha}|}
      \log \pi^{S}_{\theta}\!\left(y^{\alpha}_{t} \mid x, y^{\alpha}_{<t}\right)
    \right].
\end{equation}
For our graph-coloring $\alpha$-masked SFT experiments, we fine-tune Qwen3-4B using a global batch size of $16$ and a micro-batch size of $1$. We use 30k training steps. We optimize with AdamW at a learning rate of $1\times10^{-5}$, and no weight decay. We use a linear warmup (10 steps) followed by cosine decay to zero, and clip the gradient L2 norm at $1$. The maximum sequence length is $10{,}240$ tokens, with a maximum prompt length of $1{,}024$ tokens and a maximum response length of $8{,}192$ tokens. All SFT experiments are run on a single node with $4$ Intel Gaudi2 HPUs \cite{10.1145/3785462.3815814}.

\textbf{Note:} During inference we keep the model temperature $0.6$, top-$p=0.95$, top-$k=20$, and max response length $=32k$.
\FloatBarrier

\section{Background}\label{sec:background}
Traditionally, Knowledge distillation is a framework where there is a student-teacher pair and the student model is being trained to mimic the behaviour of the teacher by minimizing the divergence between their output distributions. This framework was first introduced by \cite{hinton2015distilling} for transferring knowledge from an ensemble or from a large highly regularized model into a smaller, distilled model. In the context of autoregressive models, knowledge distillation has been extensively studied for training a smaller Language model to mimic the outputs of a larger teacher LLM. Let ${\theta}$ denote the student model's parameters, and $\pi^{S}_{\theta}$ denote the student model's policy, differentiable w.r.t ${\theta}$. Let us denote the dataset of input-output pairs as $(X,Y)$. For a divergence $D$, we define the discrepancy between token-level distributions of $\pi^{T}$ and $\pi^{S}_{\theta}$ as
\begin{equation}
D_{\pi^{T} \parallel \pi^{S}_{\theta}}(y \mid x)
:=
\frac{1}{L_y}
\sum_{n=1}^{L_y}
D\!\left(
\pi^{T}(\cdot \mid y_{<n}, x)
\;\middle\|\;
\pi^{S}_{\theta}(\cdot \mid y_{<n}, x)
\right),
\end{equation}

There are many variants of KD studied in the literature and the distinction between the variants is based on whether the teacher's probability distribution $\pi^{T}$ is accessible during training and from where the targets, $Y$, are obtained.

\begin{itemize}
    \item \textbf{Sequence-level KD:} \cite{kim2016sequence} If $\pi^{T}$ is not accessible during training, and the outputs $Y$ are sampled from the teacher on inputs $X$, then with supervised finetuning, the student model is trained to maximize the likelihood of high-probability teacher sequences.
    \item \textbf{Supervised KD:} \cite{hinton2015distilling}, \cite{sanh2019distilbert} The student model is trained to mimic the next-token probability distribution of the teacher model. The loss $L_{\mathrm{SD}}(\theta)$ over the dataset $(X,Y)$ is calculated as 

    \begin{equation}
        \mathcal{L}_{\mathrm{SKD}}(\theta)
        :=
        \mathbb{E}_{(x,y)\sim (X,Y)}
        \left[
        D_{\mathrm{KL}}\!\left(
        \pi^{T} \;\middle\|\; \pi^{S}_{\theta}
        \right)(y \mid x)
        \right].
    \end{equation}

    \item \textbf{On-policy KD:} MinilLM \cite{gu2024minillm}, GKD \cite{agarwal2024policy}, performs on-policy knowledge distillation by generating responses from the student model.
    \begin{equation}
    \begin{split}
    \mathcal{L}_{\mathrm{GKD}}(\theta) &:=
        \mathbb{E}_{x \sim X,\; y \sim \pi^{S}_{\theta}(\cdot \mid x)} \left[
        D_{\mathrm{KL}}\!\left(
        \pi^{T}(\cdot \mid y_{<n}, x)
        \;\middle\|\;
        \pi^{S}_{\theta}(\cdot \mid y_{<n}, x)
        \right)
    \right]
    \end{split}
    \end{equation}

In \cite{agarwal2024policy}, authors have shown that on-policy KD works well with different variants of KL divergence such as forward KL, reverse KL and JSD. Sequence-level KD and Supervised KD have the problem of distribution mismatch between output sequences seen during training and those generated by the student during inference. The on-policy training in GKD addresses this drawback effectively.

    \item \textbf{Self-Distillation: } \cite{deng2023implicit, shenfeld2026self, hubotter2026reinforcement} show that the same model can be used in 2 different roles - teacher and student thus $\pi^{T} = \pi^{S}_{\theta}$. The teacher is conditioned on additional information $z$ such as in-context examples, environment feedback or CoT tokens and the student trained to mimic the teacher.

  \begin{equation}
    \mathcal{L}_{\mathrm{SD}}(\theta) :=
    \mathbb{E}_{x \sim X,\; y \sim \pi^{S}_{\theta}(\cdot \mid x)}
    \left[
      D_{\mathrm{KL}}\!\left(
        \pi^{T}(\cdot \mid y_{<n}, z, x)
        \;\middle\|\;
        \pi^{S}_{\theta}(\cdot \mid y_{<n}, x)
      \right)
    \right].
\end{equation}

\end{itemize}

\section{Dataset Details}\label{appx:dataset_detail}
\textbf{Graph Coloring}: We generate graph instances using two standard random graph algorithms to produce structurally distinct topologies. The Erd\H{o}s--R\'enyi (ER) algorithm \citep{erdds1959random} constructs a graph on $n$ vertices by including each of the $\binom{n}{2}$ possible edges independently with probability $p$. The resulting degree distribution is binomial and concentrates sharply around its mean $p(n-1)$, so ER graphs are approximately homogeneous: most vertices have a similar number of neighbors. For our in-distribution datasets, we fix the number of vertices to $n = 10$ and edges to $e = 18$ to control even the density of the graph. For generating harder instances in the {$\mathrm{GC}$-$\mathrm{ER}_{\chi_6}$, we increase the density of the graph by increasing number of edges to $e = 30$ and filter the generated instances to have a chromatic number of 6.

The Barab\'asi--Albert (BA) algorithm \citep{barabasi1999emergence} generates graphs through preferential attachment. Starting from a small seed graph, vertices are added one at a time, and each new vertex attaches $m$ edges to existing vertices with probability proportional to their current degree.  This algorithm generates heterogeneous graphs with a small number of high-degree hubs. In the {$\mathrm{GC}$-$\mathrm{BA}_{\chi_6}$ test set, we have fixed the number of vertices to $n = 10$ and edges to $e = 30$ and filter the generated instances to have a chromatic number of 6.  All graphs are generated using NetworkX \citep{SciPyProceedings_11}.

\textbf{Countdown}: This domain \citep{wikipediaCountdowngame} is a generalized version of the classic 24 Game \citep{yang2022chain}, where the objective is to combine a set of input numbers using basic arithmetic operations (+, –, ×, ÷) to reach a specified target number. In our ID dataset, each problem consists of 3 to 4 two-digit input numbers, with the target number also being a two-digit number. The dataset contains 8.5K examples, split into 8K training instances and 500 test instances. For OOD testing, we have three datasets, each perturbing a different axis. \emph{Target-range shift} keeps the input cardinality and digit count fixed but shifts the target to a one-digit number ($500$ problems). \emph{Search-depth shift} keeps the digit counts fixed but increases the input cardinality to $5$ numbers respectively ($500$ problems).

\section{Additional Results}\label{Appx:additional_results}
\subsection{Countdown Results}
\label{app:countdown-results}
\begin{figure}[!ht]
    \centering
    \includegraphics[width=0.9\textwidth]{figures/legends/regular_masking_legend_horizontal.png}
    
    \begin{subfigure}{0.33\textwidth}
        \centering
        \includegraphics[width=\textwidth]{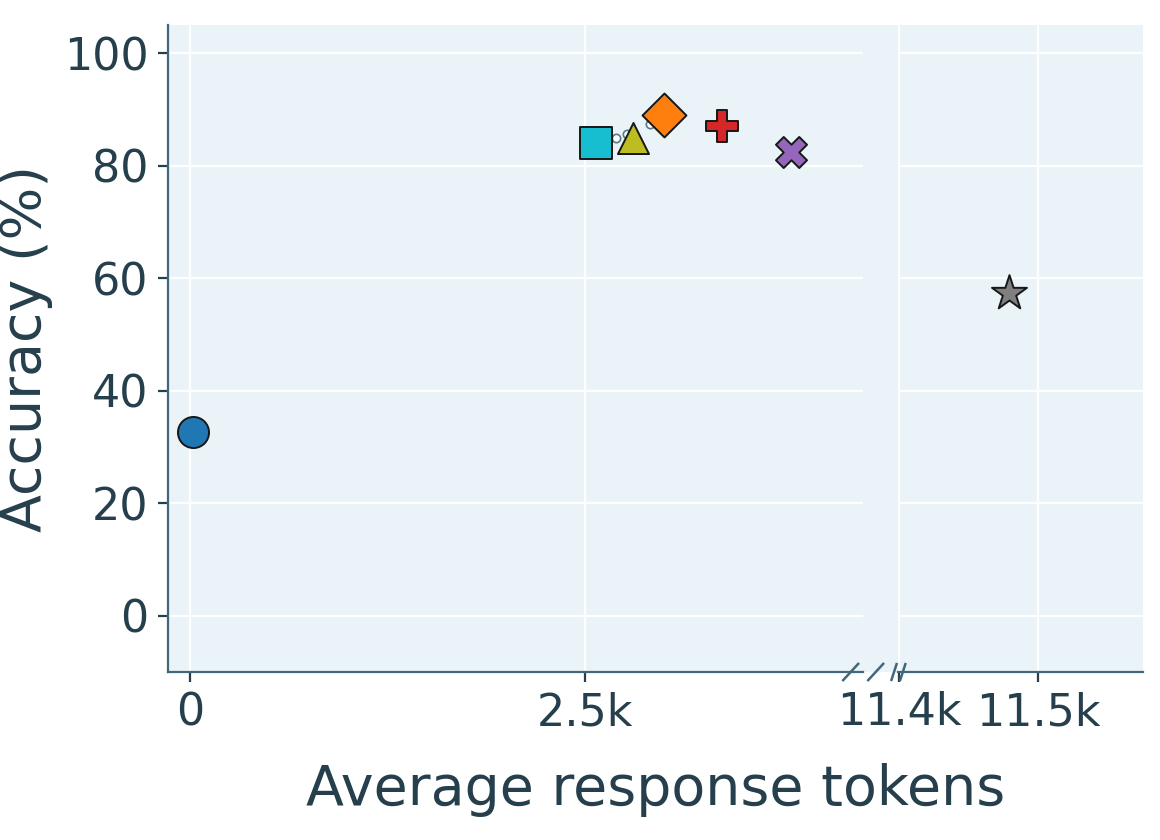}
        \caption{ID}
        \label{fig:pareto-ctdn-4b}
    \end{subfigure}\hfill
    \begin{subfigure}{0.33\textwidth}
        \centering
        \includegraphics[width=\textwidth]{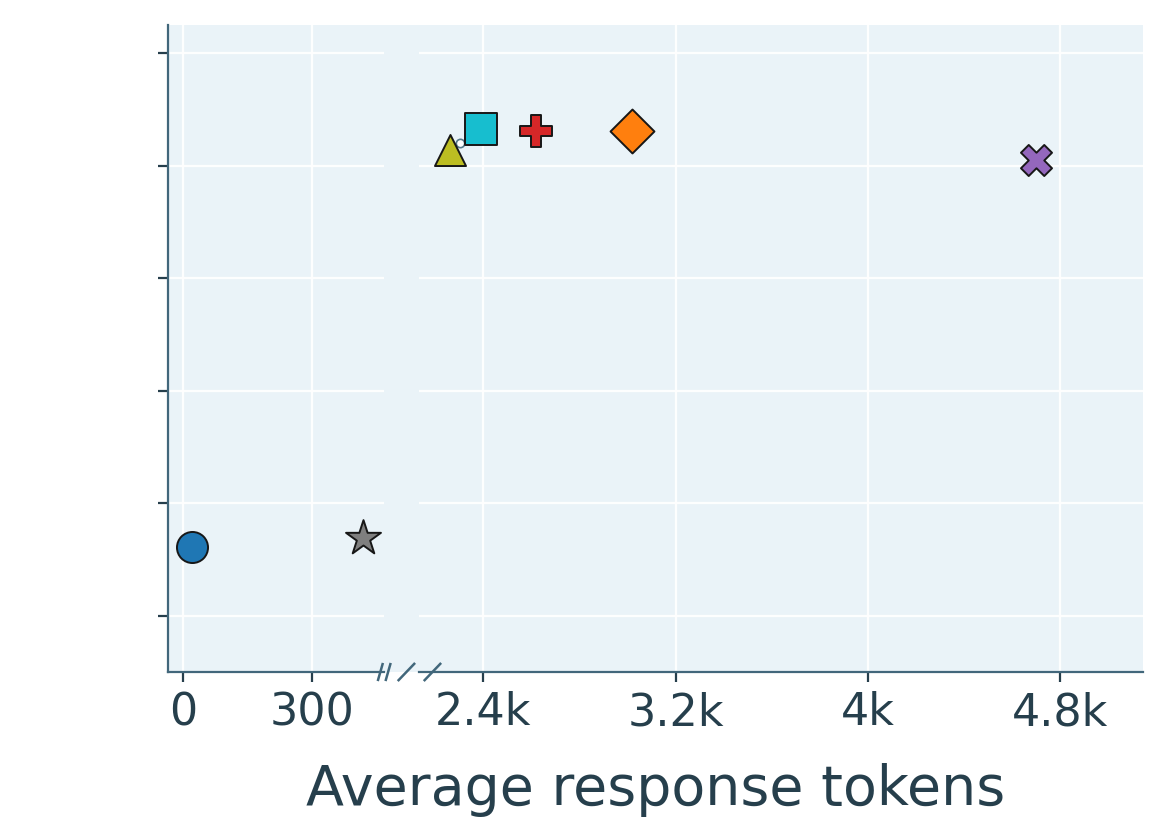}
        \caption{Target-range shift}
        \label{fig:pareto-target_3to4-4b}
    \end{subfigure}\hfill
    \begin{subfigure}{0.33\textwidth}
        \centering
        \includegraphics[width=\textwidth]{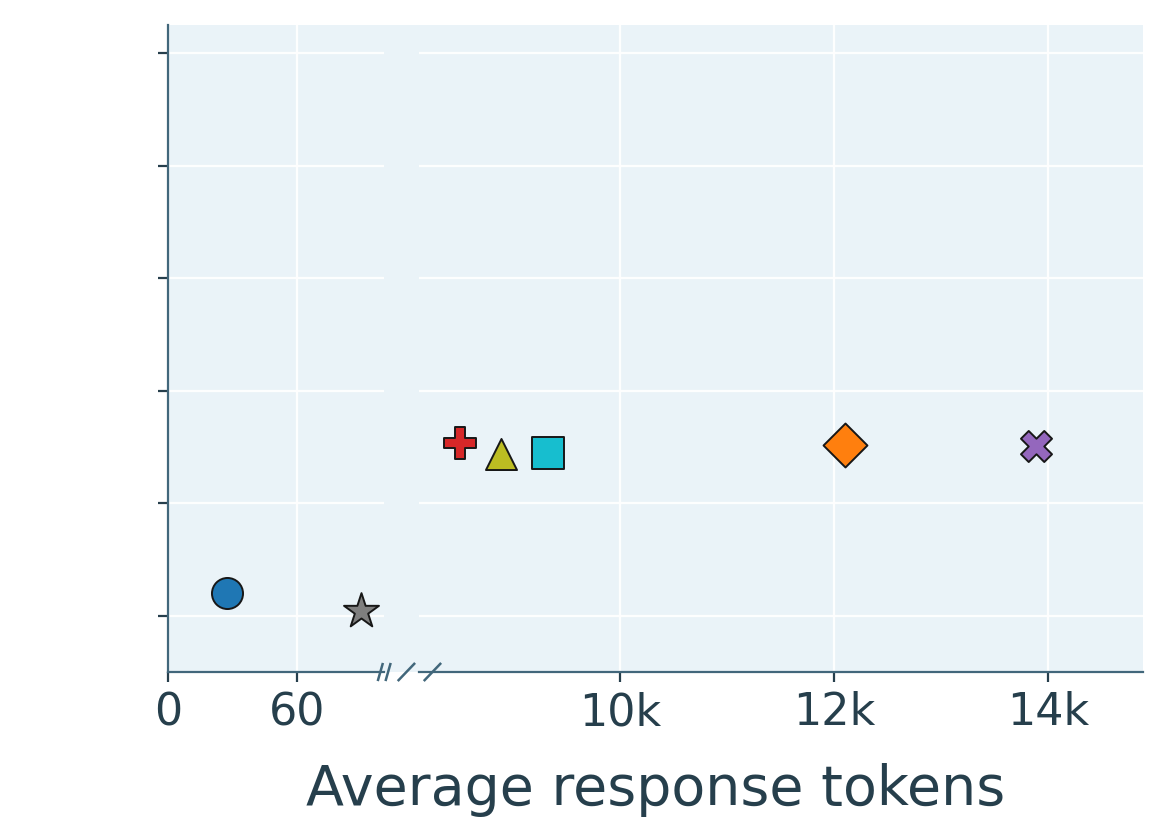}
        \caption{Search-depth shift}
        \label{fig:pareto-input5-4b}
    \end{subfigure}
    \caption{\textit{Qwen3-4B Pareto plot}: accuracy vs.\ mean response length comparison on countdown domain test datasets for $\alpha$-masked variants, teacher (base think) and student (base no-think).}
    \label{fig:pareto-self-distill-4b-countdown}    
\end{figure}

Figure~\ref{fig:pareto-self-distill-4b-countdown} shows the Pareto plot for Qwen3-4B on the countdown test sets for all baselines. The $\alpha$-masked variants ($\alpha \in \{0.3, 0.5, 0.7\}$) reach the same level of accuracy as the teacher on every countdown test set. On the in-distribution and target-shift sets, however, they save only a few hundred tokens relative to the teacher, and on the search-depth-shift set they produce more tokens than the teacher. This may be due to the mode-seeking behavior of the reverse-KL objective discussed in Section~\ref{subsec:internalization}.


\subsection{Qwen3-4B Results}\label{app:4b_multi_seed}
Tables~\ref{tab:qwen3-4b-math-32k-seeds} and~\ref{tab:qwen3-4b-gc-32k-seeds} report the accuracy and mean response length for all $\alpha$-masked variants, together with the base student and teacher models, on the math and graph-coloring test datasets, respectively.

\begin{table}[!ht]
\centering
\footnotesize
\setlength{\tabcolsep}{4pt}
\begin{tabular}{lccccccc}
\toprule
& \multicolumn{5}{c}{Masked fraction} & \multicolumn{2}{c}{Base} \\
\cmidrule(lr){2-6}\cmidrule(lr){7-8}
Dataset & 100\% & 70\% & 50\% & 30\% & 0\% & think & no-think \\
\midrule

GSM8K & 91.4$_{\pm 0.5}$ & 92.9$_{\pm 0.4}$ & 92.8$_{\pm 0.2}$ & 92.7$_{\pm 0.5}$ & 93.5$_{\pm 0.5}$ & \textbf{94.2}$_{\pm 0.1}$ & 86.1$_{\pm 0.4}$ \\
      & {\scriptsize 302$_{\pm 3}$} & {\scriptsize 1124$_{\pm 12}$} & {\scriptsize 1536$_{\pm 71}$} & {\scriptsize 1961$_{\pm 78}$} & {\scriptsize 1061$_{\pm 79}$} & {\scriptsize 2332$_{\pm 27}$} & {\scriptsize \underline{258}$_{\pm 2}$} \\[0.35em]

MATH500 & 74.1$_{\pm 1.0}$ & 75.9$_{\pm 1.0}$ & 75.7$_{\pm 0.1}$ & 74.9$_{\pm 1.1}$ & \textbf{78.9}$_{\pm 0.1}$ & 76.9$_{\pm 0.3}$ & 75.2$_{\pm 0.7}$ \\
        & {\scriptsize \underline{763}$_{\pm 29}$} & {\scriptsize 3080$_{\pm 125}$} & {\scriptsize 3671$_{\pm 61}$} & {\scriptsize 4722$_{\pm 557}$} & {\scriptsize 3340$_{\pm 146}$} & {\scriptsize 5137$_{\pm 214}$} & {\scriptsize 1051$_{\pm 34}$} \\[0.35em]

AIME25 & 17.8$_{\pm 1.9}$ & 44.4$_{\pm 1.9}$ & 51.1$_{\pm 1.9}$ & 54.4$_{\pm 6.9}$ & 46.7$_{\pm 3.3}$ & \textbf{61.1}$_{\pm 5.1}$ & 18.9$_{\pm 1.9}$ \\
       & {\scriptsize \underline{1609}$_{\pm 116}$} & {\scriptsize 13146$_{\pm 1500}$} & {\scriptsize 13730$_{\pm 1206}$} & {\scriptsize 14673$_{\pm 154}$} & {\scriptsize 12598$_{\pm 2081}$} & {\scriptsize 18190$_{\pm 840}$} & {\scriptsize 5552$_{\pm 1560}$} \\

\bottomrule
\end{tabular}
\caption{\textit{Qwen3-4B, math domain}: accuracy and mean token length for each masked fraction, trained on GSM8K. Each entry is the mean over 3 random seeds, with the sample standard deviation in the subscript. Accuracy (\%) is on the first line; the smaller line is mean response length in tokens. Bold marks the highest accuracy in each row; underline marks the shortest mean response.}
\label{tab:qwen3-4b-math-32k-seeds}
\end{table}

\begin{table}[!ht]
\centering
\footnotesize
\setlength{\tabcolsep}{4pt}
\begin{tabular}{lccccccc}
\toprule
& \multicolumn{5}{c}{Masked fraction} & \multicolumn{2}{c}{Base} \\
\cmidrule(lr){2-6}\cmidrule(lr){7-8}
Dataset & 100\% & 70\% & 50\% & 30\% & 0\% & think & no-think \\
\midrule

$\mathrm{GC}$-$\mathrm{ER}_{\chi_{3\text{-}4}}$ & 11.5$_{\pm 0.3}$ & \textbf{97.5}$_{\pm 1.0}$ & 97.0$_{\pm 0.7}$ & 95.8$_{\pm 1.1}$ & 92.3$_{\pm 0.6}$ & 91.9$_{\pm 0.6}$ & 29.2$_{\pm 1.4}$ \\
 & {\scriptsize \underline{67}$_{\pm 0}$} & {\scriptsize 3274$_{\pm 129}$} & {\scriptsize 3772$_{\pm 115}$} & {\scriptsize 4112$_{\pm 129}$} & {\scriptsize 6369$_{\pm 120}$} & {\scriptsize 6185$_{\pm 237}$} & {\scriptsize 1576$_{\pm 85}$} \\[0.35em]

$\mathrm{GC}$-$\mathrm{ER}_{\chi_{6}}$ & 10.5$_{\pm 0.2}$ & \textbf{79.5}$_{\pm 1.5}$ & 75.5$_{\pm 0.7}$ & 71.1$_{\pm 1.5}$ & 46.7$_{\pm 1.6}$ & 45.4$_{\pm 2.5}$ & 13.7$_{\pm 2.1}$ \\
 & {\scriptsize \underline{66}$_{\pm 3}$} & {\scriptsize 6906$_{\pm 196}$} & {\scriptsize 7986$_{\pm 389}$} & {\scriptsize 9384$_{\pm 344}$} & {\scriptsize 12439$_{\pm 487}$} & {\scriptsize 15489$_{\pm 643}$} & {\scriptsize 1254$_{\pm 166}$} \\[0.35em]

$\mathrm{GC}$-$\mathrm{BA}_{\chi_{6}}$ & 10.6$_{\pm 0.3}$ & 88.3$_{\pm 1.0}$ & \textbf{89.9}$_{\pm 1.3}$ & 87.3$_{\pm 1.9}$ & 65.7$_{\pm 1.8}$ & 62.1$_{\pm 2.6}$ & 18.4$_{\pm 1.9}$ \\
 & {\scriptsize \underline{115}$_{\pm 36}$} & {\scriptsize 5185$_{\pm 135}$} & {\scriptsize 5410$_{\pm 152}$} & {\scriptsize 5942$_{\pm 382}$} & {\scriptsize 10186$_{\pm 366}$} & {\scriptsize 12134$_{\pm 298}$} & {\scriptsize 1566$_{\pm 445}$} \\

\bottomrule
\end{tabular}
\caption{\textit{Qwen3-4B, graph coloring domain}: accuracy and mean token length for each masked fraction, trained on $\mathrm{GC}$-$\mathrm{ER}_{\chi_{3\text{-}4}}$. Each entry is the mean over 3 random seeds, with the sample standard deviation in the subscript. Accuracy (\%) is on the first line; the smaller line is mean response length in tokens. Bold marks the highest accuracy in each row; underline marks the shortest mean response.}
\label{tab:qwen3-4b-gc-32k-seeds}
\end{table}

\FloatBarrier
\subsection{Qwen3-8B Results}
\label{app:8b_multi_seed}
Tables~\ref{tab:qwen3-8b-math-32k-seeds} and~\ref{tab:qwen3-8b-gc-32k-seeds} report the accuracy and mean response length for all $\alpha$-masked variants, together with the base student and teacher models, on the math and graph-coloring test datasets, respectively.

\begin{table}[!ht]
\centering
\footnotesize
\setlength{\tabcolsep}{4pt}
\begin{tabular}{lccccccc}
\toprule
& \multicolumn{5}{c}{Masked fraction} & \multicolumn{2}{c}{Base} \\
\cmidrule(lr){2-6}\cmidrule(lr){7-8}
Dataset & 100\% & 70\% & 50\% & 30\% & 0\% & think & no-think \\
\midrule

GSM8K & 92.8$_{\pm 0.1}$ & 94.0$_{\pm 0.2}$ & 94.5$_{\pm 0.1}$ & 93.9$_{\pm 0.5}$ & \textbf{95.0}$_{\pm 0.3}$ & 94.6$_{\pm 0.2}$ & 80.7$_{\pm 0.4}$ \\
      & {\scriptsize 302$_{\pm 1}$} & {\scriptsize 1429$_{\pm 30}$} & {\scriptsize 1884$_{\pm 45}$} & {\scriptsize 2141$_{\pm 91}$} & {\scriptsize 1109$_{\pm 12}$} & {\scriptsize 2402$_{\pm 6}$} & {\scriptsize \underline{266}$_{\pm 2}$} \\[0.35em]

MATH500 & 74.9$_{\pm 0.2}$ & 73.5$_{\pm 0.8}$ & 76.3$_{\pm 0.4}$ & 76.3$_{\pm 0.1}$ & 73.4$_{\pm 1.4}$ & \textbf{77.5}$_{\pm 0.1}$ & 73.8$_{\pm 2.5}$ \\
        & {\scriptsize \underline{805}$_{\pm 37}$} & {\scriptsize 4472$_{\pm 365}$} & {\scriptsize 4249$_{\pm 77}$} & {\scriptsize 4619$_{\pm 42}$} & {\scriptsize 3348$_{\pm 64}$} & {\scriptsize 5355$_{\pm 116}$} & {\scriptsize 1181$_{\pm 120}$} \\[0.35em]

AIME25 & 21.1$_{\pm 1.9}$ & 45.6$_{\pm 8.4}$ & 56.7$_{\pm 3.3}$ & \textbf{62.2}$_{\pm 3.8}$ & 41.1$_{\pm 13.5}$ & 61.1$_{\pm 1.9}$ & 18.9$_{\pm 1.9}$ \\
       & {\scriptsize \underline{1848}$_{\pm 561}$} & {\scriptsize 13587$_{\pm 746}$} & {\scriptsize 14989$_{\pm 1333}$} & {\scriptsize 16322$_{\pm 297}$} & {\scriptsize 10924$_{\pm 1271}$} & {\scriptsize 17274$_{\pm 752}$} & {\scriptsize 4121$_{\pm 1077}$} \\

\bottomrule
\end{tabular}
\caption{\textit{Qwen3-8B, math domain}: accuracy and mean token length for each masked fraction, trained on GSM8K. Each entry is the mean over 3 random seeds, with the sample standard deviation in the subscript. Accuracy (\%) is on the first line; the smaller line is mean response length in tokens. Bold marks the highest accuracy in each row; underline marks the shortest mean response.}
\label{tab:qwen3-8b-math-32k-seeds}
\end{table}

\begin{table}[!ht]
\centering
\footnotesize
\setlength{\tabcolsep}{4pt}
\begin{tabular}{lccccccc}
\toprule
& \multicolumn{5}{c}{Masked fraction} & \multicolumn{2}{c}{Base} \\
\cmidrule(lr){2-6}\cmidrule(lr){7-8}
Dataset & 100\% & 70\% & 50\% & 30\% & 0\% & think & no-think \\
\midrule

$\mathrm{GC}$-$\mathrm{ER}_{\chi_{3\text{-}4}}$ & 22.1$_{\pm 0.5}$ & \textbf{98.5}$_{\pm 0.6}$ & \textbf{98.5}$_{\pm 0.8}$ & 97.6$_{\pm 0.5}$ & 94.3$_{\pm 1.0}$ & 95.5$_{\pm 0.3}$ & 33.2$_{\pm 1.8}$ \\
 & {\scriptsize \underline{62}$_{\pm 0}$} & {\scriptsize 3650$_{\pm 123}$} & {\scriptsize 3766$_{\pm 233}$} & {\scriptsize 4277$_{\pm 196}$} & {\scriptsize 5253$_{\pm 69}$} & {\scriptsize 5752$_{\pm 131}$} & {\scriptsize 1149$_{\pm 28}$} \\[0.35em]

$\mathrm{GC}$-$\mathrm{ER}_{\chi_{6}}$ & 13.0$_{\pm 0.1}$ & 80.4$_{\pm 1.0}$ & \textbf{85.6}$_{\pm 0.6}$ & 78.3$_{\pm 2.2}$ & 54.4$_{\pm 1.9}$ & 64.4$_{\pm 2.5}$ & 13.5$_{\pm 0.5}$ \\
 & {\scriptsize \underline{211}$_{\pm 25}$} & {\scriptsize 9598$_{\pm 405}$} & {\scriptsize 8289$_{\pm 181}$} & {\scriptsize 10224$_{\pm 691}$} & {\scriptsize 11855$_{\pm 266}$} & {\scriptsize 14434$_{\pm 12}$} & {\scriptsize 1217$_{\pm 23}$} \\[0.35em]

$\mathrm{GC}$-$\mathrm{BA}_{\chi_{6}}$ & 15.2$_{\pm 0.9}$ & 91.4$_{\pm 0.9}$ & \textbf{93.3}$_{\pm 1.0}$ & 91.3$_{\pm 0.4}$ & 76.5$_{\pm 1.9}$ & 81.5$_{\pm 0.4}$ & 18.4$_{\pm 1.9}$ \\
 & {\scriptsize \underline{198}$_{\pm 27}$} & {\scriptsize 6304$_{\pm 227}$} & {\scriptsize 5509$_{\pm 164}$} & {\scriptsize 6148$_{\pm 241}$} & {\scriptsize 8470$_{\pm 290}$} & {\scriptsize 10298$_{\pm 288}$} & {\scriptsize 1145$_{\pm 22}$} \\

\bottomrule
\end{tabular}
\caption{\textit{Qwen3-8B, graph coloring domain}: accuracy and mean token length for each masked fraction, trained on $\mathrm{GC}$-$\mathrm{ER}_{\chi_{3\text{-}4}}$. Each entry is the mean over 3 random seeds, with the sample standard deviation in the subscript. Accuracy (\%) is on the first line; the smaller line is mean response length in tokens. Bold marks the highest accuracy in each row; underline marks the shortest mean response.}
\label{tab:qwen3-8b-gc-32k-seeds}
\end{table}

\FloatBarrier
\subsection{Out-of-Domain Qwen3-4B Results}

\begin{figure}[!ht]
    \centering

    \includegraphics[width=0.8\textwidth]{figures/legends/cross_domain_masking_legend_horizontal.png}

    \vspace{0.5em}
    \begin{subfigure}[t]{0.33\textwidth}
        \centering
        \includegraphics[width=\textwidth]{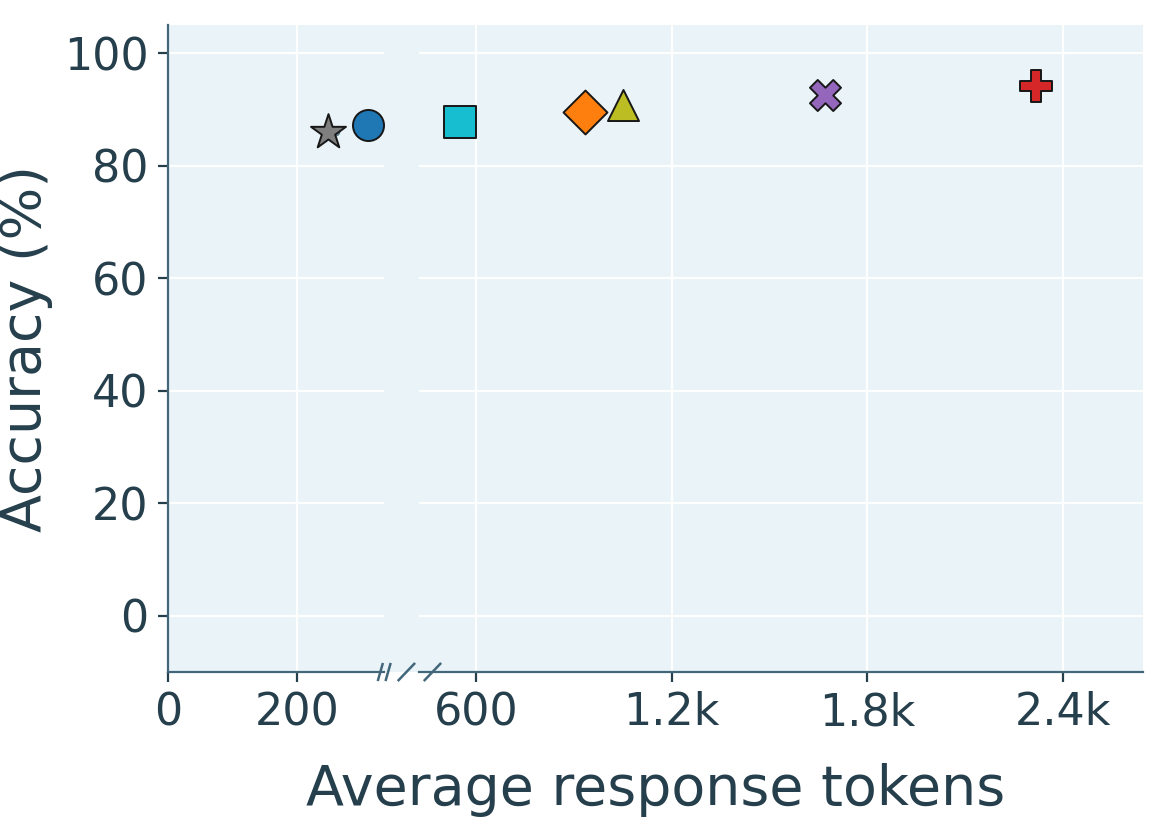}
    \end{subfigure}%
    \hspace{0.05\textwidth}%
    \begin{subfigure}[t]{0.33\textwidth}
        \centering
        \includegraphics[width=\textwidth]{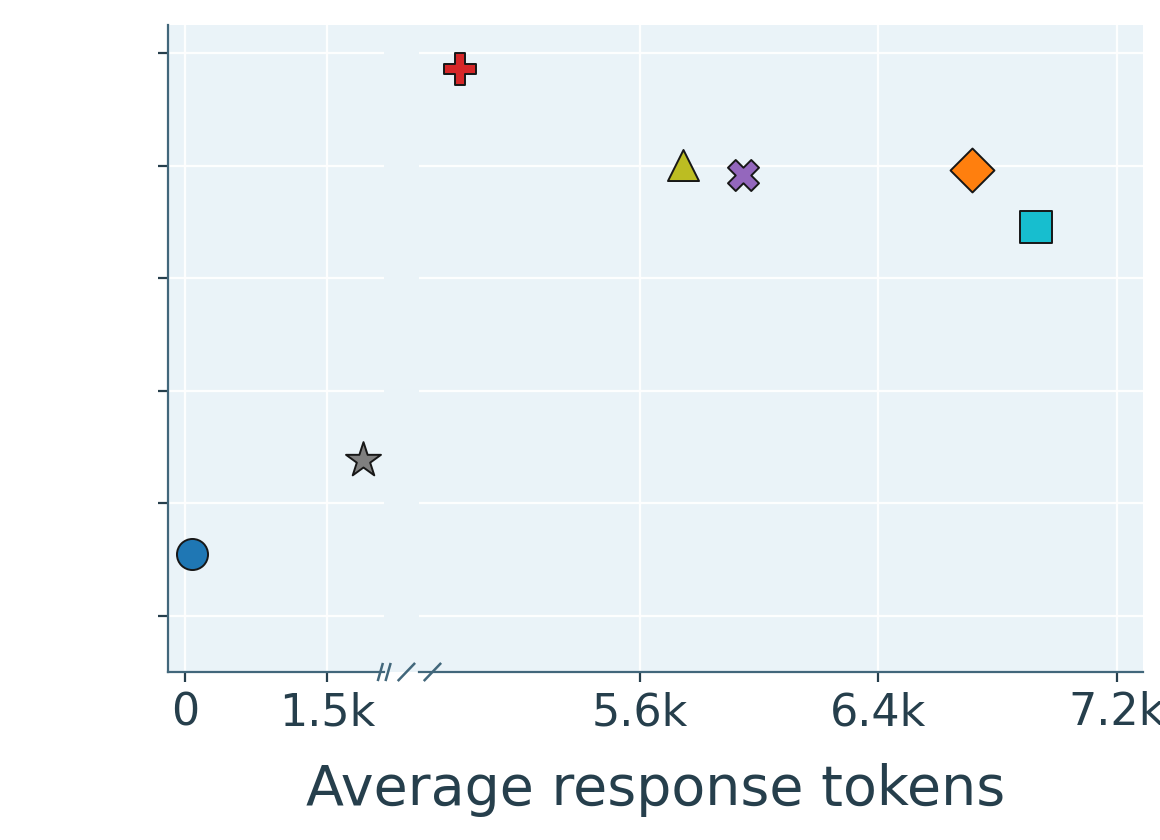}
    \end{subfigure}

    \caption{\textit{Out-of-domain Qwen3-4B Pareto plots:} (left) graph coloring models evaluated on the GSM8K test set and (right) GSM8K models evaluated on $\mathrm{GC}$-$\mathrm{ER}_{\chi_{3-4}}$ test set. }
    \label{fig:cross-domain-4b}
\end{figure}

Figure~\ref{fig:cross-domain-4b} shows out-of-domain evaluation for Qwen3-4B models, together with the base student and teacher model. The results are pretty similar to the Qwen3-8B out-of-domain results. The models trained on the graph coloring domain retain their inference cost savings even on the GSM8K test dataset. However, the models trained on the GSM8K dataset have slightly lower performance on the $\mathrm{GC}$-$\mathrm{ER}_{\chi_{3-4}}$ test set compared to the base teacher model and also produce more tokens.

\FloatBarrier
\section{Response length distributions}\label{subsec:length-dist}

In the Pareto plots, to compare the inference efficiency of each baseline, we use the mean response length. However, the mean alone does not provide the full picture of how the model's response lengths are distributed across the entire test set. A high mean response length can arise either because the model consistently generates long responses across the test set or because it generates very long responses for only a small fraction of problems. In other words, the response-length distribution can exhibit either unimodal or bimodal behavior. To examine this, we plot violin plots, which combine a box plot with a kernel density estimate and therefore show the overall shape of the response-length distribution.

\begin{wrapfigure}{r}{0.42\textwidth}
    \centering
    \includegraphics[width=0.4\textwidth]{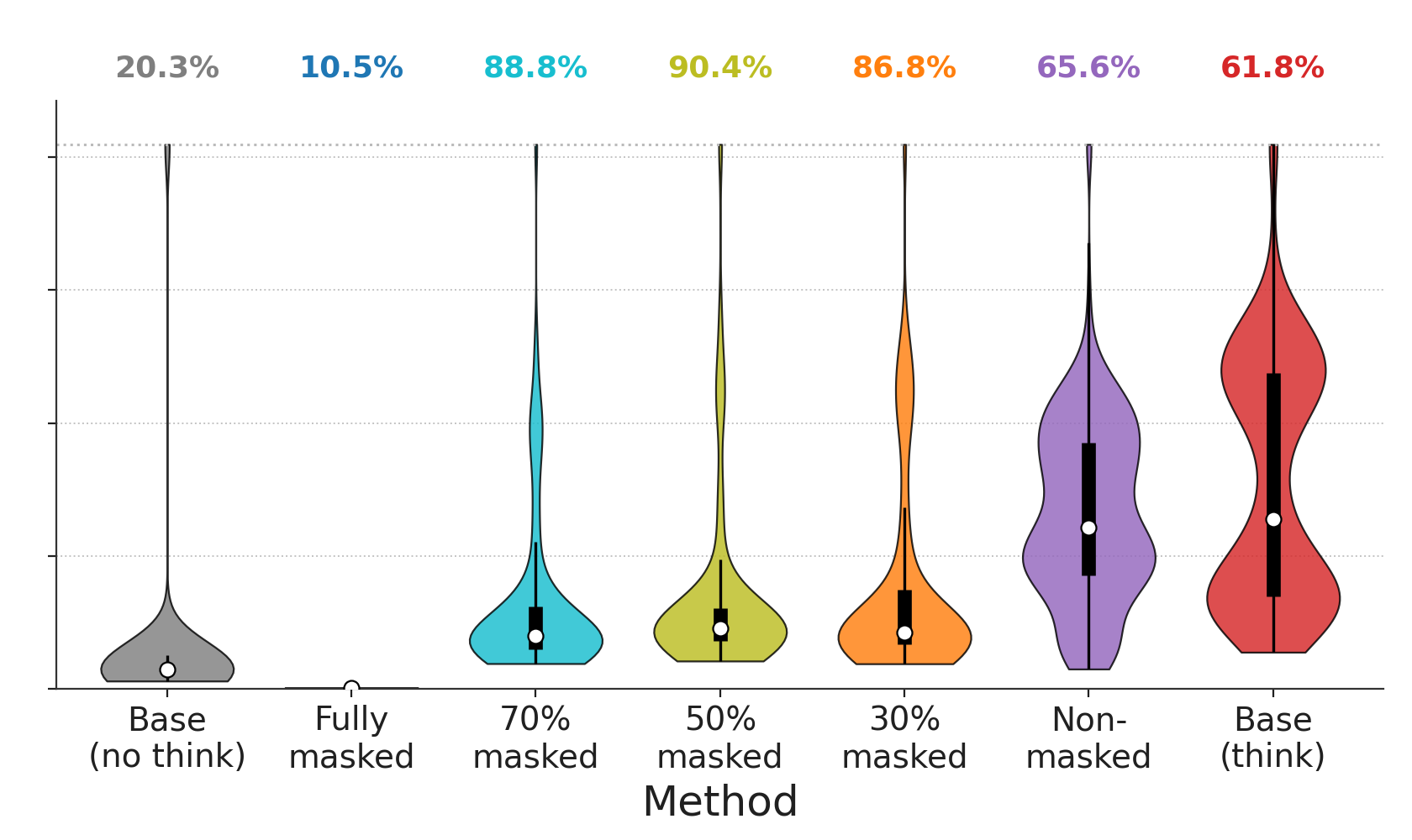}
    \caption{Violin Plot}
    \label{fig:violin-gc-er-id-4b-8k}
\end{wrapfigure}

Figure~\ref{fig:violin-gc-er-id-4b-8k} shows the violin plots for all baselines on the graph-coloring OOD-BA test set under a maximum token budget of 32k. From the figure, we can see that the teacher and the non-masked variant exhibit clear bimodal response-length distributions and produce responses with a median length of approximately 10k tokens. In contrast, the $\alpha$-masked variants ($\alpha \in {0.3,0.5,0.7}$) exhibit a largely uni-modal distribution with a median response length of approximately 3k tokens. This suggests that, by internalizing a fraction of the reasoning trace, the masked variants learn to generate shorter responses more consistently across the test set. Figures~\ref{fig:violin-self-distill-4b}  and \ref{fig:violin-self-distill-8b} show the violin plots for all domain test datasets for Qwen3-4B and Qwen3-8B respectively.

\begin{figure}[!ht]
    \centering
    \includegraphics[width=0.9\textwidth]{figures/legends/regular_masking_legend_horizontal.png}
    
    \begin{subfigure}{0.33\textwidth}
        \centering
        \includegraphics[width=\textwidth]{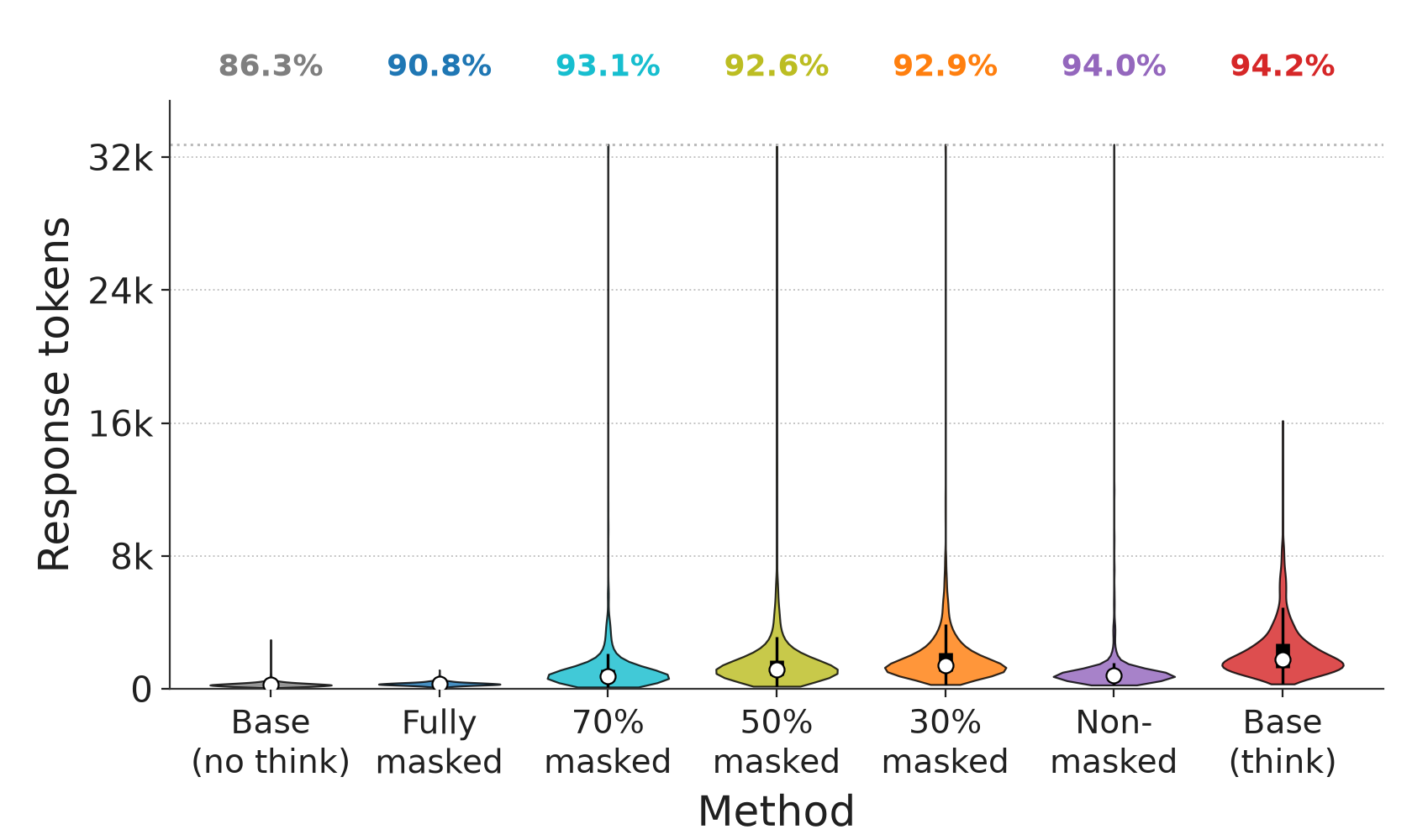}
        \caption{GSM8K}
        \label{fig:violin-gsm8k-4b}
    \end{subfigure}\hfill
    \begin{subfigure}{0.33\textwidth}
        \centering
        \includegraphics[width=\textwidth]{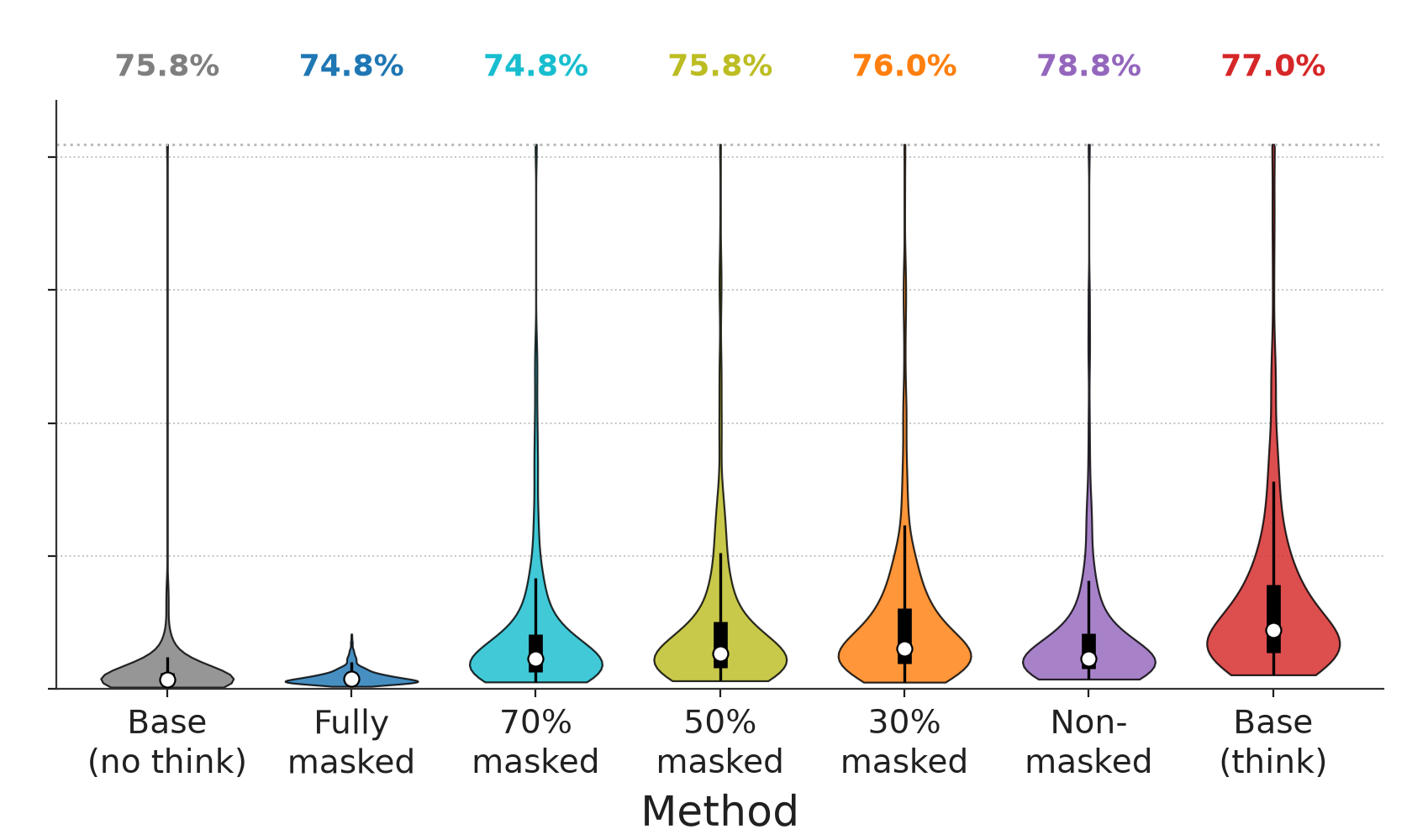}
        \caption{MATH-500}
        \label{fig:violin-math-4b}
    \end{subfigure}\hfill
    \begin{subfigure}{0.33\textwidth}
        \centering
        \includegraphics[width=\textwidth]{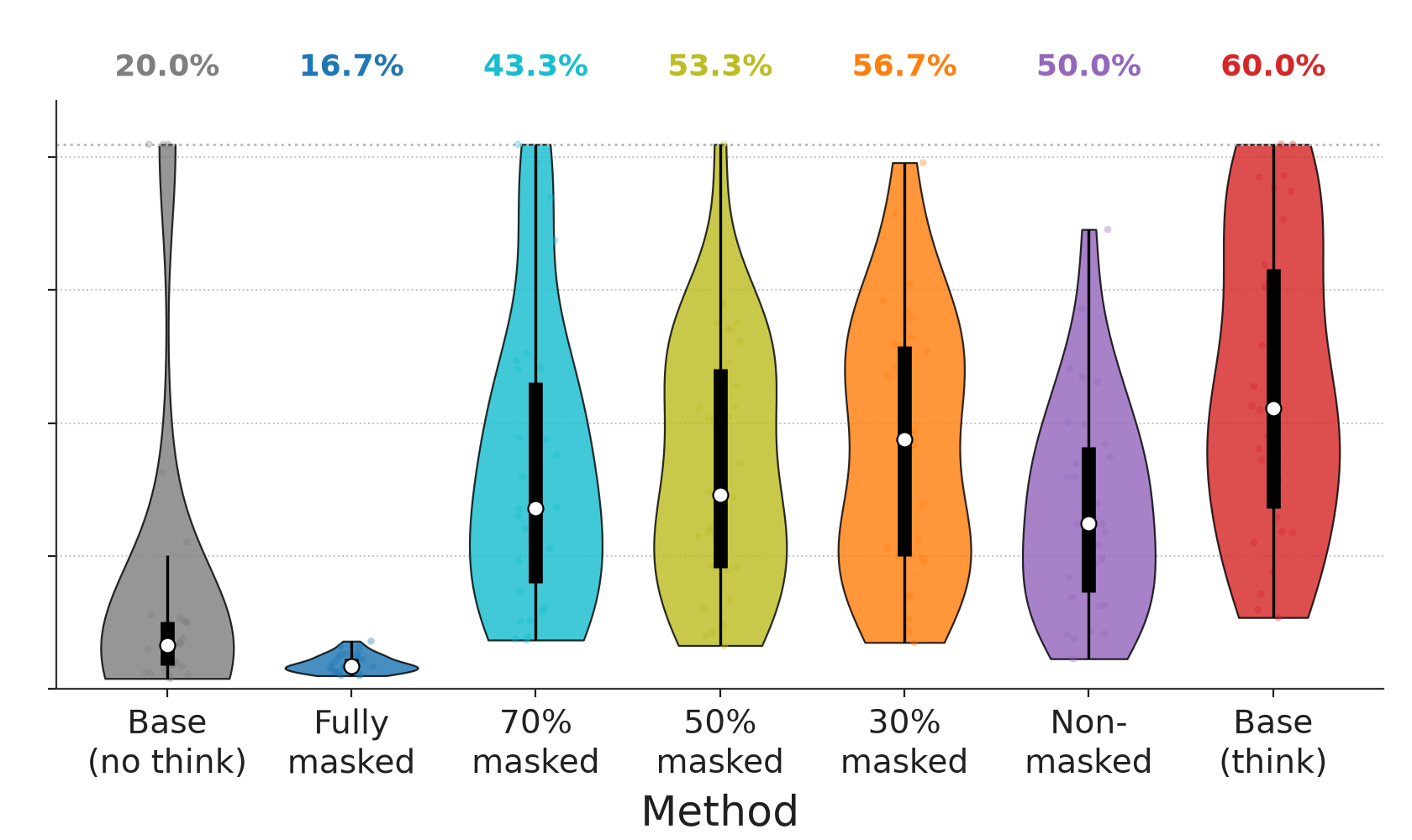}
        \caption{AIME25}
        \label{fig:violin-aime-4b}
    \end{subfigure}
    
    \begin{subfigure}{0.33\textwidth}
        \centering
        \includegraphics[width=\textwidth]{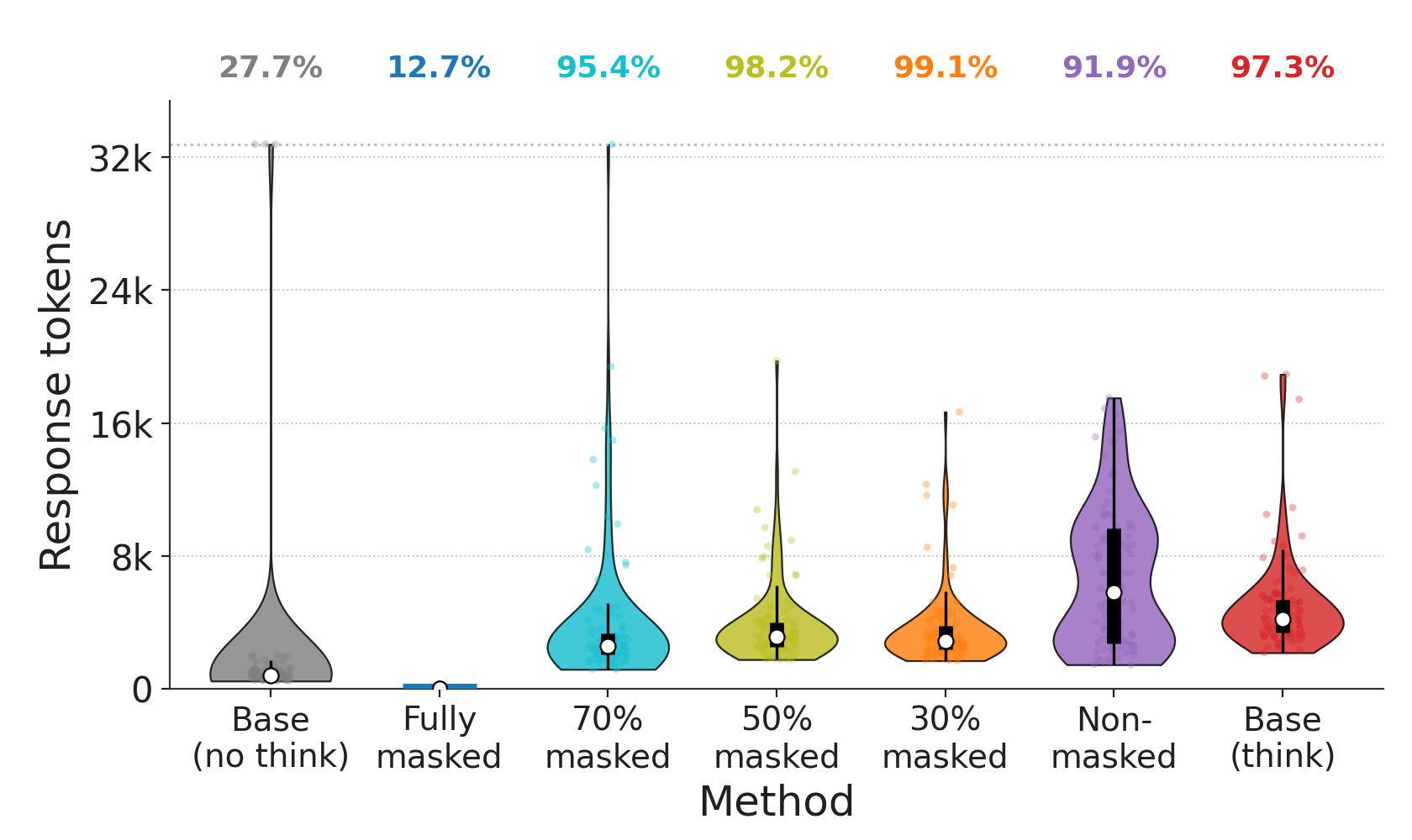}
        \caption{$\mathrm{GC}$-$\mathrm{ER}_{\chi_{3-4}}$}
        \label{fig:violin-gc-er-id-4b}
    \end{subfigure}\hfill
    \begin{subfigure}{0.33\textwidth}
        \centering
        \includegraphics[width=\textwidth]{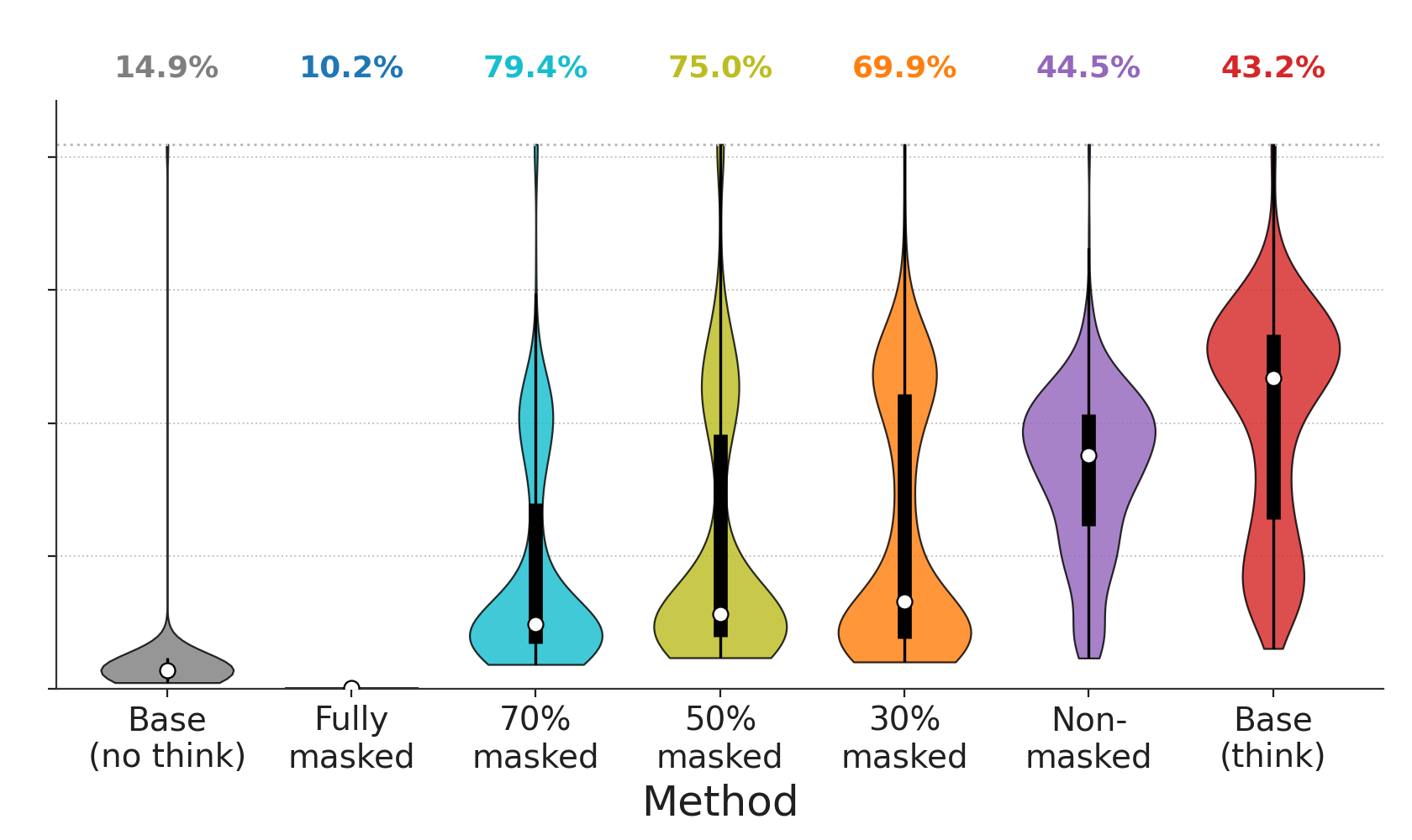}
        \caption{$\mathrm{GC}$-$\mathrm{ER}_{\chi_6}$}
        \label{fig:violin-gc-er-ood-4b}
    \end{subfigure}\hfill
    \begin{subfigure}{0.33\textwidth}
        \centering
        \includegraphics[width=\textwidth]{figures/qwen3_4b_panels/length_dist_pooled/graph_coloring/panels/violin/cut/violin_ood_ba_v10_m30_nonplanar_r32k.png}
        \caption{$\mathrm{GC}$-$\mathrm{BA}_{\chi_6}$}
        \label{fig:violin-gc-ba-ood-4b}
    \end{subfigure}

    
    \caption{\textit{Qwen3-4B Violin plot}: Token distribution comparison on math (top row), graph-coloring (middle row) and countdown (bottom row) domain test datasets for $\alpha$-masked variants, teacher (base think) and student (base no-think).}
    \label{fig:violin-self-distill-4b}    
\end{figure}

\begin{figure}[!ht]
    \centering
    \includegraphics[width=0.9\textwidth]{figures/legends/regular_masking_legend_horizontal.png}
    
    \begin{subfigure}{0.33\textwidth}
        \centering
        \includegraphics[width=\textwidth]{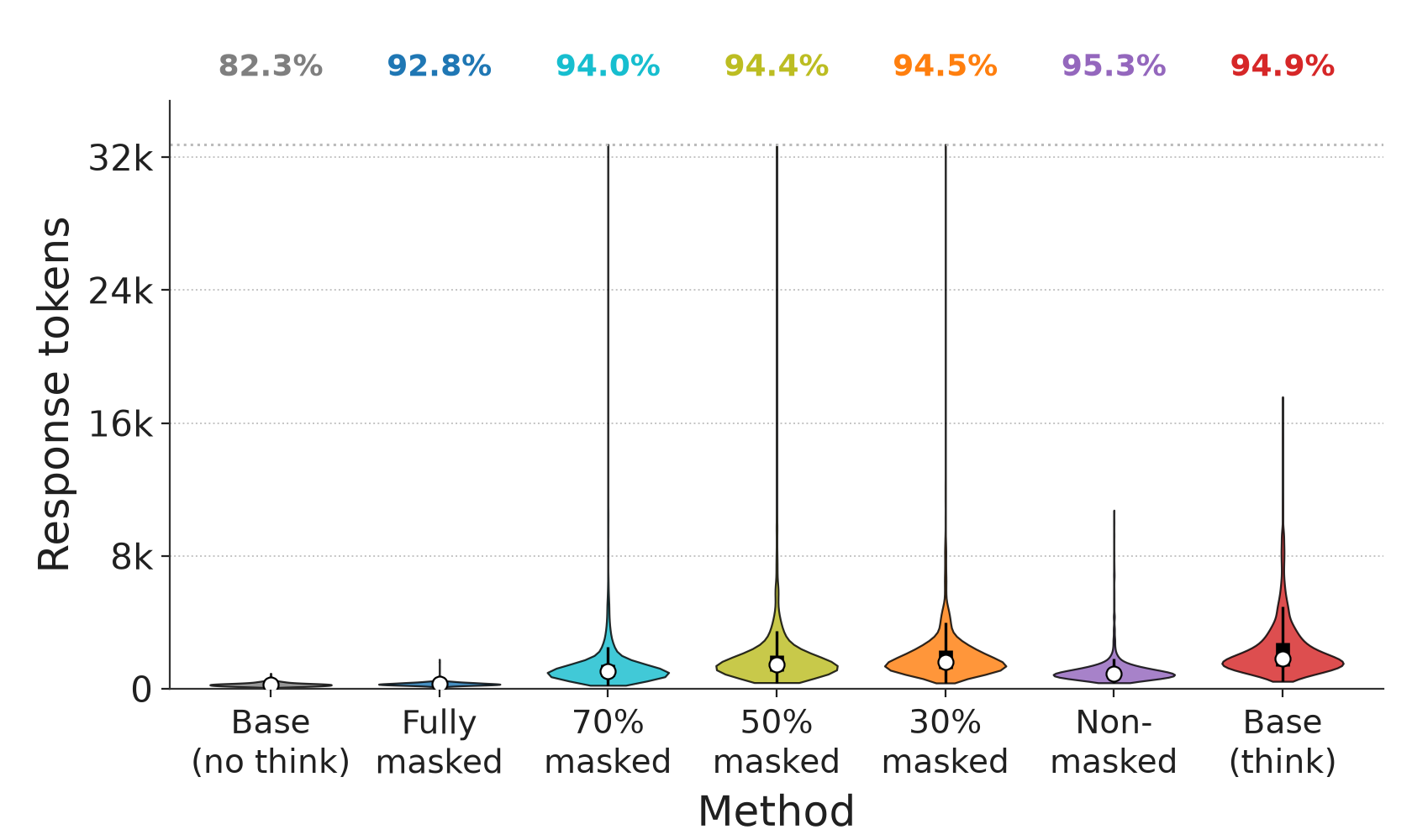}
        \caption{GSM8K}
        \label{fig:violin-gsm8k-8b}
    \end{subfigure}\hfill
    \begin{subfigure}{0.33\textwidth}
        \centering
        \includegraphics[width=\textwidth]{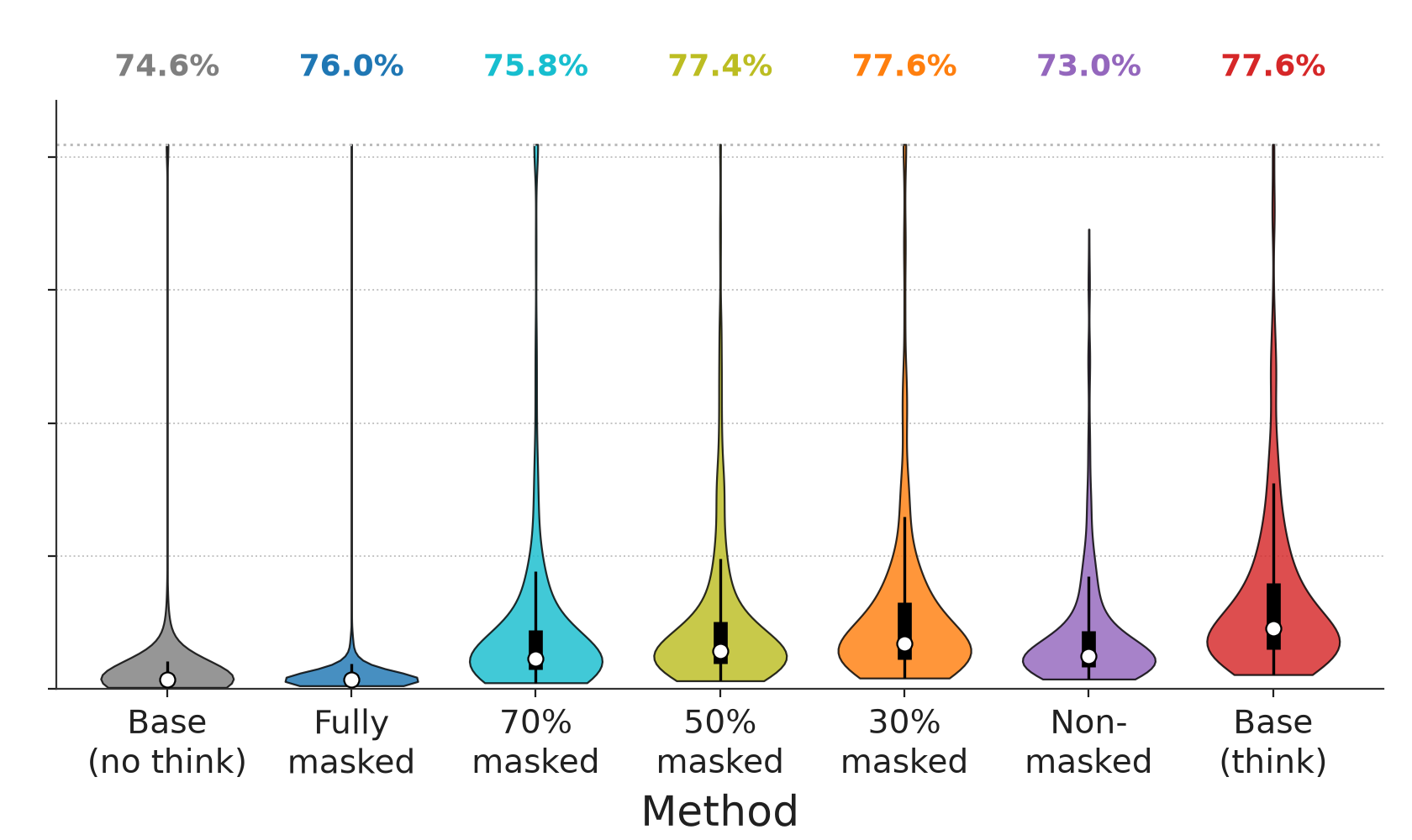}
        \caption{MATH-500}
        \label{fig:violin-math-8b}
    \end{subfigure}\hfill
    \begin{subfigure}{0.33\textwidth}
        \centering
        \includegraphics[width=\textwidth]{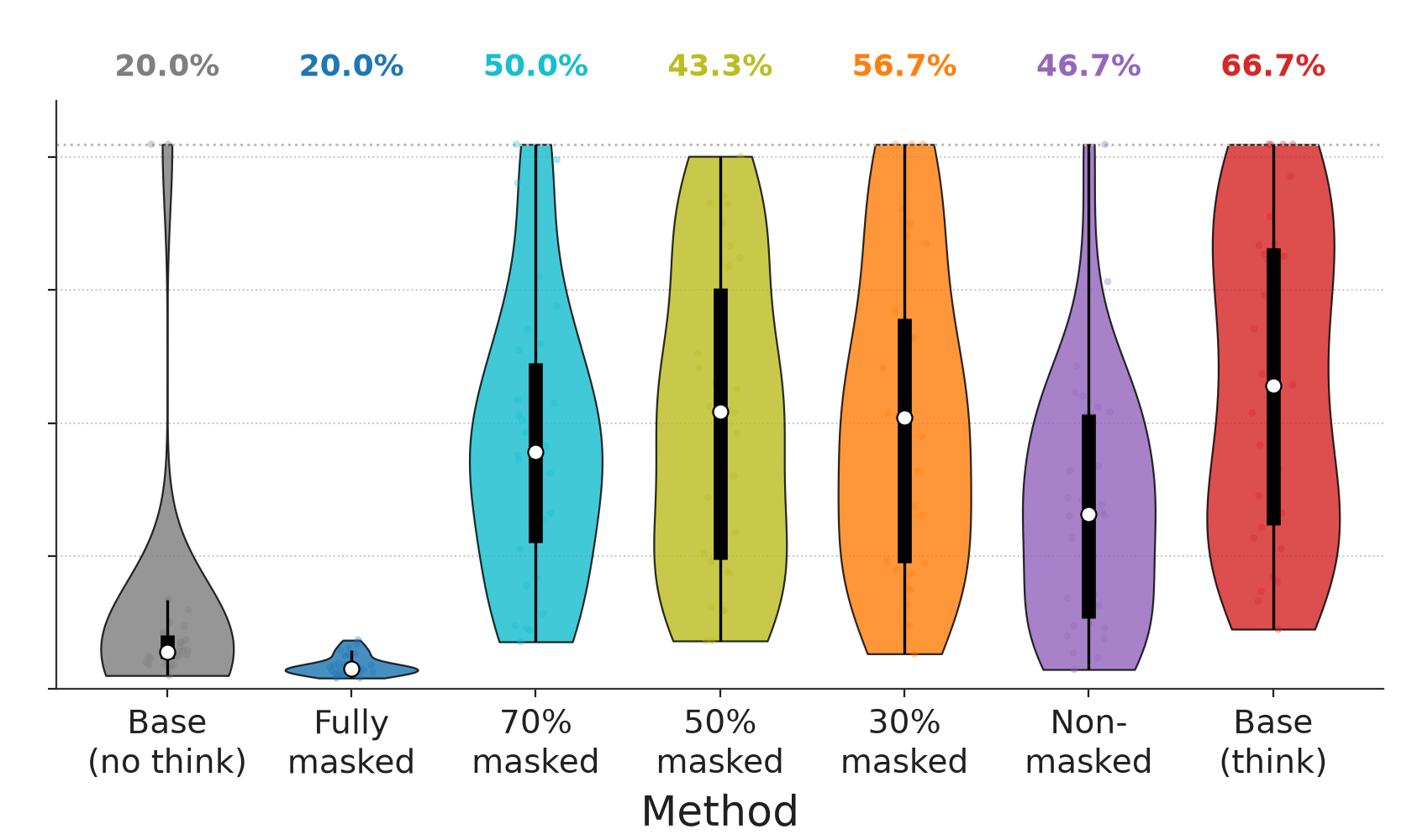}
        \caption{AIME25}
        \label{fig:violin-aime-8b}
    \end{subfigure}
    
    \begin{subfigure}{0.33\textwidth}
        \centering
        \includegraphics[width=\textwidth]{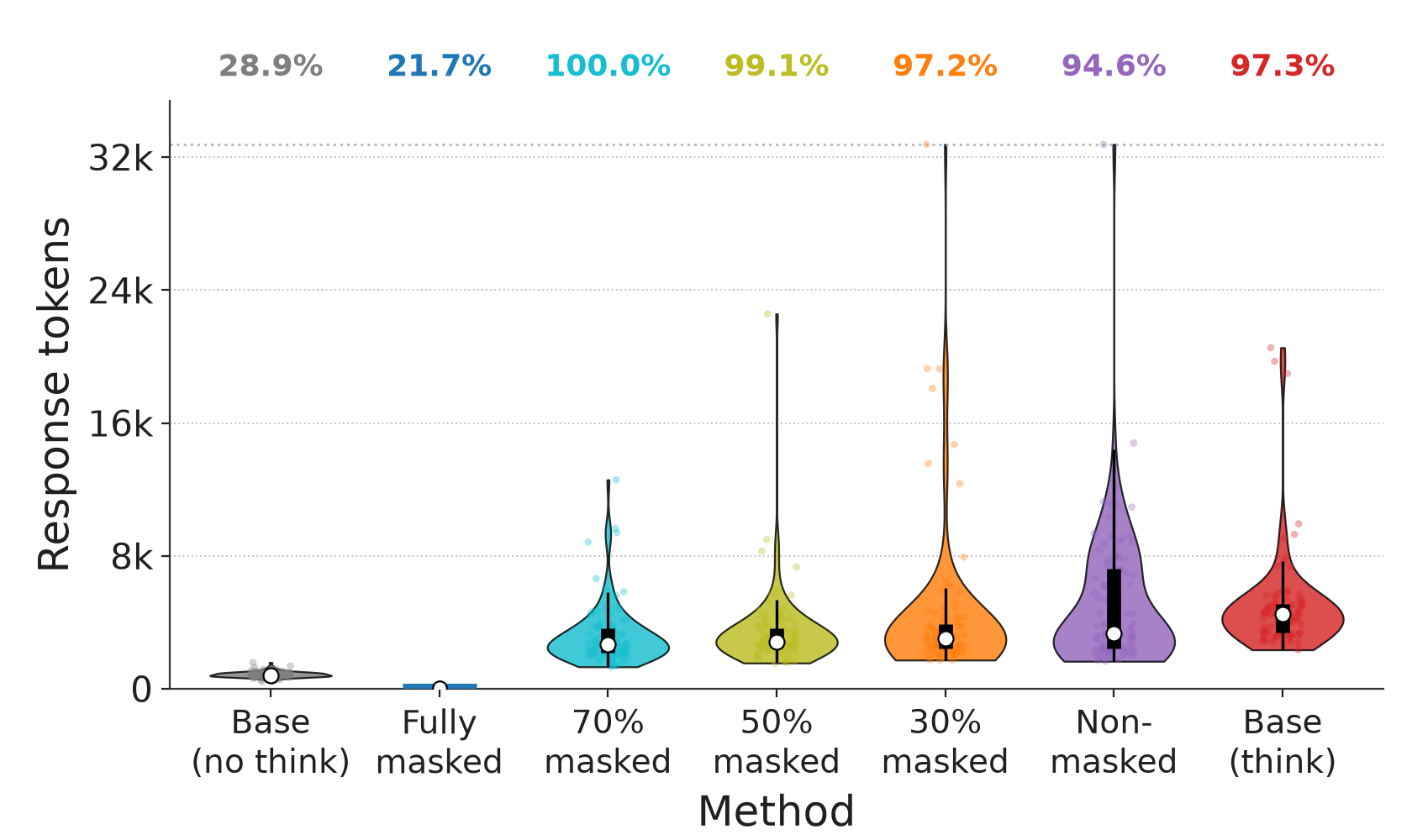}
        \caption{$\mathrm{GC}$-$\mathrm{ER}_{\chi_{3-4}}$}
        \label{fig:violin-gc-er-id-8b}
    \end{subfigure}\hfill
    \begin{subfigure}{0.33\textwidth}
        \centering
        \includegraphics[width=\textwidth]{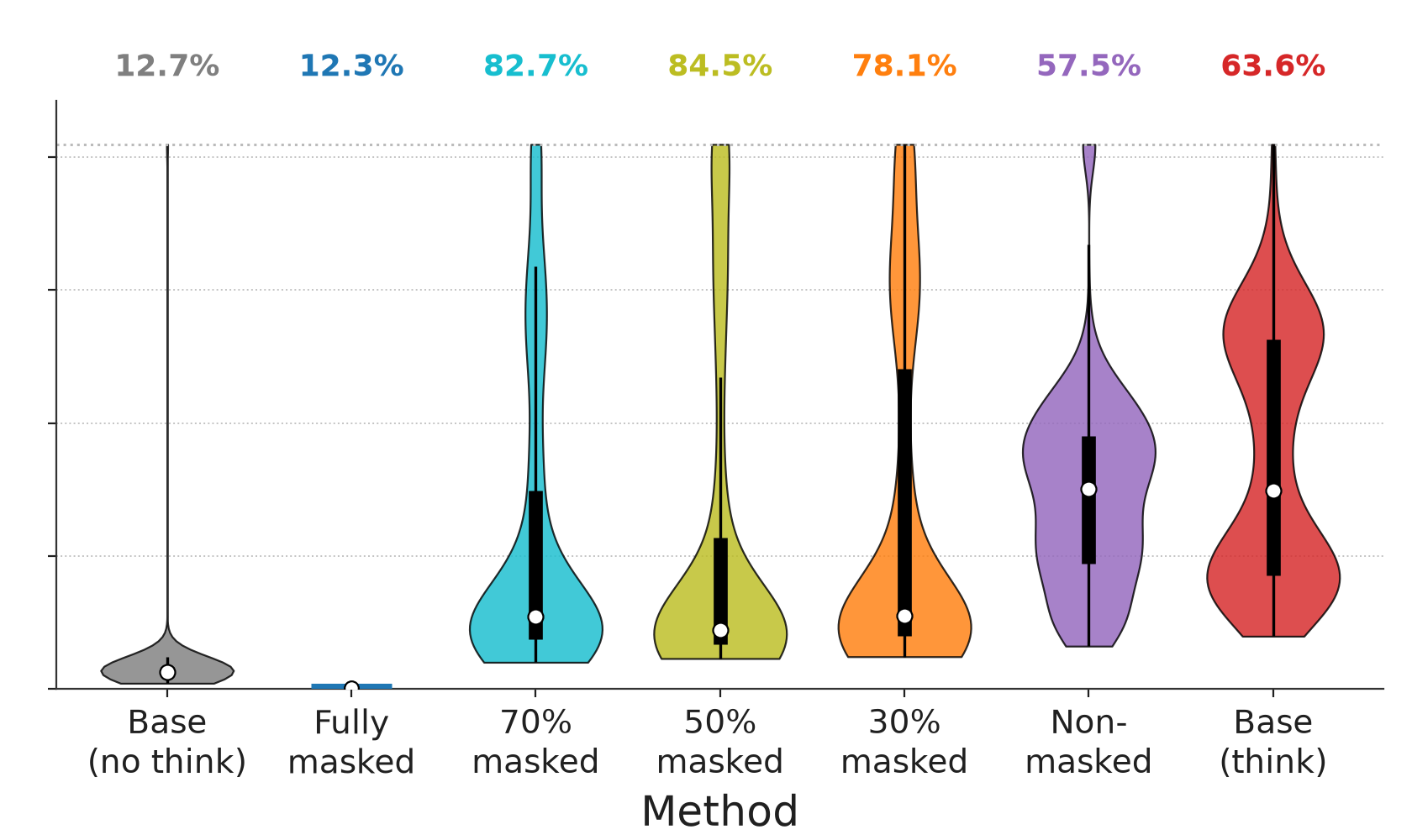}
        \caption{$\mathrm{GC}$-$\mathrm{ER}_{\chi_6}$}
        \label{fig:violin-gc-er-ood-8b}
    \end{subfigure}\hfill
    \begin{subfigure}{0.33\textwidth}
        \centering
        \includegraphics[width=\textwidth]{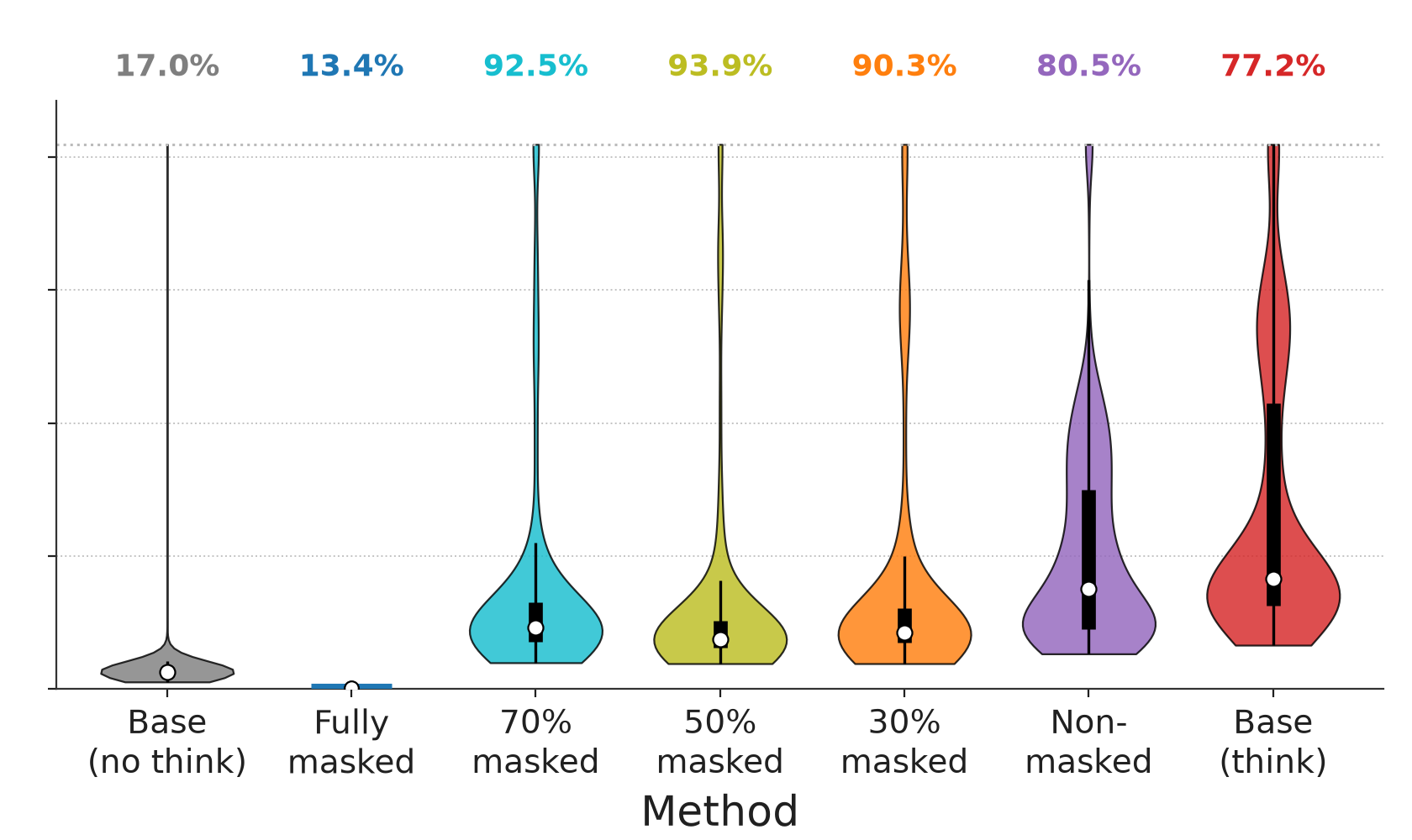}
        \caption{$\mathrm{GC}$-$\mathrm{BA}_{\chi_6}$}
        \label{fig:violin-gc-ba-ood-8b}
    \end{subfigure}

    
    \caption{\textit{Qwen3-8B Violin plot}: Token distribution comparison on math (top row), graph-coloring (middle row) and countdown (bottom row) domain test datasets for $\alpha$-masked variants, teacher (base think) and student (base no-think).}
    \label{fig:violin-self-distill-8b}    
\end{figure}
\FloatBarrier
\section{Prompt format and the responses of masked self distilled models}
\label{app:trace-eg}
We show a $\mathrm{GC}$-$\mathrm{ER}_{\chi_{3-4}}$ test instance prompt and responses of every masked self-distilled variant. We can see that the non-masked variant produces a response that is very similar to the teacher. However, the responses of models trained with  $\alpha \in \{0.3, 0.5, 0.7\}$ seem disconnected, as if beginning from an intermediate point in the reasoning rather than from the start.

\begin{figure}[H]
    \centering
    \includegraphics[width=0.82\linewidth]{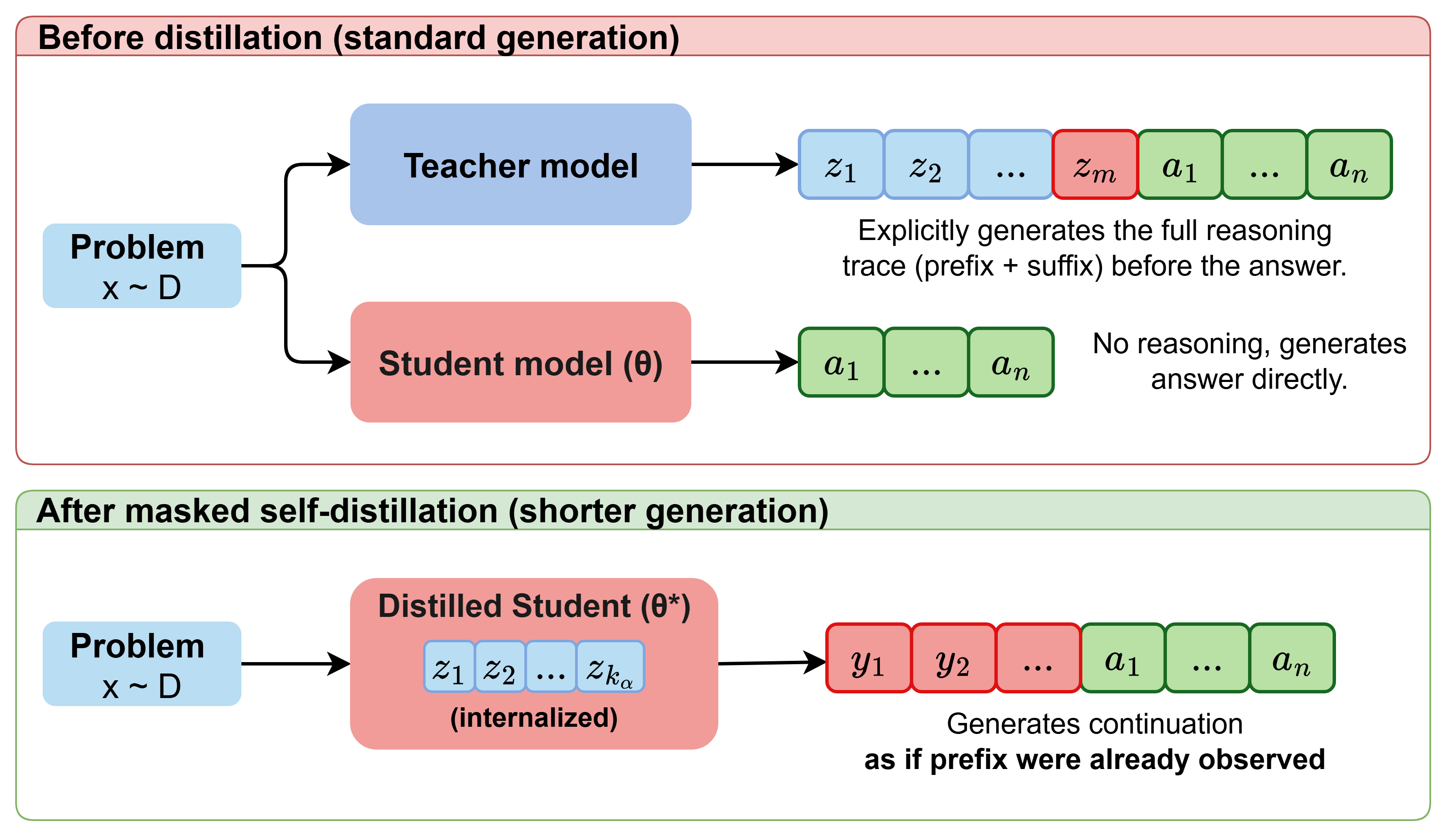}
    \caption{Illustration of standard generation before distillation and shorter generation after masked self-distillation.}
    \label{fig:masked-self-distillation-generation}
\end{figure}

\begin{tcolorbox}[
    colback=gray!5, colframe=black!60, boxrule=0.5pt,
    fonttitle=\bfseries, title={Prompt},
    breakable,
    width=\linewidth,
    boxsep=1mm, left=1mm, right=1mm, top=1mm, bottom=1mm
]
\begin{Verbatim}[fontsize=\scriptsize]
Please provide a coloring for this graph such that every vertex is not connected to a 
vertex of the same color. The graph has these properties:

Vertices: [0, 1, 2, 3, 4, 5, 6, 7, 8, 9]
Edges: [(0, 1), (0, 7), (0, 9), (1, 2), (1, 3), (1, 5), (1, 7), (1, 8), (1, 9), (2, 3), 
(2, 4), (2, 8), (3, 4), (3, 6), (3, 8), (3, 9), (5, 6), (5, 9)]
Possible colors: [1, 2, 3, 4]

Return your final answer in <answer> </answer> tags as a JSON map of vertices to colors,
 for example <answer> {"0": 1, "1": 2, "2": 3} </answer>.
assistant

\end{Verbatim}
\end{tcolorbox}
\vspace{0.5em}









\begin{tcolorbox}[
    colback=gray!5, colframe=black!60, boxrule=0.5pt,
    fonttitle=\bfseries, title={Qwen3-8B base, think (Teacher) response},
    breakable,
    width=\linewidth,
    boxsep=1mm, left=1mm, right=1mm, top=1mm, bottom=1mm
]
\begin{Verbatim}[fontsize=\scriptsize]
<think>
Okay, let's see. I need to color this graph with four colors so that no two connected 
vertices share the same color. The vertices are numbered 0 through 9, and there are a 
bunch of edges listed. Let me start by understanding the graph structure.

First, I should probably draw the graph or at least note down which vertices are 
connected to which. Let me list out the edges again:

Edges:
(0,1), (0,7), (0,9),
(1,2), (1,3), (1,5), (1,7), (1,8), (1,9),
(2,3), (2,4), (2,8),
(3,4), (3,6), (3,8), (3,9),
(5,6), (5,9)

So vertex 0 is connected to 1,7,9. Vertex 1 is connected to 0,2,3,5,7,8,9. That's a lot.
 Vertex 2 is connected to 1,3,4,8. Vertex 3 is connected to 1,2,4,6,8,9. Vertex 4 is 
conn

[... 7421 characters of reasoning elided ...]

ct. Vertex 2 is color 1, neighbors are 1 (2), 3 (3), 4 (2), 8 (4). Correct. Vertex 7 is 
color 3, neighbors are 0 (1), 1 (2). Correct. Vertex 1 is color 2, neighbors are 0 (1), 
2 (1), 3 (3), 5 (1), 7 (3), 8 (4), 9 (4). All different. Vertex 0 is color 1, neighbors 
are 1 (2), 7 (3), 9 (4). All different.

Yes, this seems to work. So the answer should be this coloring.
</think>

<answer>
{
  "0": 1,
  "1": 2,
  "2": 1,
  "3": 3,
  "4": 2,
  "5": 1,
  "6": 2,
  "7": 3,
  "8": 4,
  "9": 4
}
</answer>
\end{Verbatim}
\end{tcolorbox}

\begin{tcolorbox}[
    colback=gray!5, colframe=black!60, boxrule=0.5pt,
    fonttitle=\bfseries, title={Non-Masked variant response},
    breakable,
    width=\linewidth,
    boxsep=1mm, left=1mm, right=1mm, top=1mm, bottom=1mm
]
\begin{Verbatim}[fontsize=\scriptsize]
<think>
Okay, let's tackle this graph coloring problem. The goal is to assign colors from 
[1,2,3,4] to each vertex such that no two connected vertices share the same color. The 
graph has vertices 0-9 and specific edges. Let me first try to understand the structure 
of the graph.

First, I'll list out all the edges again to make sure I have them right:

Edges:
(0,1), (0,7), (0,9),
(1,2), (1,3), (1,5), (1,7), (1,8), (1,9),
(2,3), (2,4), (2,8),
(3,4), (3,6), (3,8), (3,9),
(5,6), (5,9)

So, each vertex is connected to others as per these edges. Let me try to visualize or 
note the connections for each vertex.

Starting with vertex 0: connected to 1,7,9.
Vertex 1 is connected to 0,2,3,5,7,8,9. That

[... 4027 characters of reasoning elided ...]

), 1 (2), 3 (3), 5 (1). All different. 

So this seems to work. Let me check if any conflicts were missed. For example, vertex 3 
is connected to 9 (color 4) - yes, 3 is color 3. Vertex 5 is connected to 9 (color 4) - 
5 is color 1. All okay.

So the color assignments are:

0:1, 1:2, 2:1, 3:3, 4:2, 5:1, 6:2, 7:3, 8:4, 9:4.

Which matches the correct solution provided. So that's the answer.
</think>

<answer>
{"0": 1, "1": 2, "2": 1, "3": 3, "4": 2, "5": 1, "6": 2, "7": 3, "8": 4, "9": 4}
</answer>
\end{Verbatim}
\end{tcolorbox}

\begin{tcolorbox}[
    colback=gray!5, colframe=black!60, boxrule=0.5pt,
    fonttitle=\bfseries, title={30\%-Masked variant response},
    breakable,
    width=\linewidth,
    boxsep=1mm, left=1mm, right=1mm, top=1mm, bottom=1mm
]
\begin{Verbatim}[fontsize=\scriptsize]
 1, 3, 5, 7, 8, 9. Wait, vertex 1 is connected to 0,2,3,5,7,8,9. So vertex 1's neighbors
 are 0,2,3,5,7,8,9. So vertex 1 is connected to 0,2,3,5,7,8,9. So vertex 1 is connected 
to 0, which is color 1. So vertex 1's color can't be 1. Wait, but vertex 1's color is 1?
 Wait, no, I'm trying to assign colors. Let me think again.

Wait, maybe I need to start with vertex 0, then vertex 1, etc. Let me try to go step by 
step.

Let me try to color vertex 0 first. Let's assign color 1 to vertex 0. Then, vertices 
connected to 0 (1,7,9) can't be color 1.

Next, vertex 1. It's connected to 0 (color 1), so it can't be color 1. Let's assign 
color 2 to vertex 1. Then, its neighbors (0,2,3,5,7,8,9) can't be col

[... 5654 characters of reasoning elided ...]

3 and 2 -> different. Edge (3,6): 3 and 2 -> different. Edge (3,8): 3 and 4 -> 
different. Edge (3,9): 3 and 4 -> different. Edge (2,3): 1 and 3 -> different. Edge 
(2,4): 1 and 2 -> different. Edge (2,8): 1 and 4 -> different. Edge (5,6): 1 and 2 -> 
different. Edge (5,9): 1 and 4 -> different. All edges are okay.

So this seems to be a valid coloring. Therefore, the answer should be this mapping.
</think>

<answer>
{"0": 1, "1": 2, "2": 1, "3": 3, "4": 2, "5": 1, "6": 2, "7": 3, "8": 4, "9": 4}
</answer>
\end{Verbatim}
\end{tcolorbox}

\begin{tcolorbox}[
    colback=gray!5, colframe=black!60, boxrule=0.5pt,
    fonttitle=\bfseries, title={50\%-Masked variant response},
    breakable,
    width=\linewidth,
    boxsep=1mm, left=1mm, right=1mm, top=1mm, bottom=1mm
]
\begin{Verbatim}[fontsize=\scriptsize]
 1, 2, 3, 4, 5, 6, 7, 8, 9. Wait, vertex 1 is color 2, vertex 3 is color 3, vertex 5 is 
color 1? Wait, no. Wait, vertex 5 is connected to 1 (color 1?), no. Wait, vertex 1 is 
color 2. Wait, I need to track all the colors.

Wait, let me retrace. Let me list all vertices with their current colors:

Vertex 0: 1

Vertex 1: 2

Vertex 2: 1

Vertex 3: 3

Vertex 4: 2

Vertex 5: ?

Wait, I think I need to go step by step again. Maybe I made a mistake in the order.

Let me try again, more carefully.

Start with vertex 0. Assign color 1.

Vertex 0: 1

Now, vertex 1 is connected to 0. So vertex 1 can't be 1. Assign color 2.

Vertex 1: 2

Now, vertex 2 is connected to 1. So vertex 2 can't be 2. Assign col

[... 3053 characters of reasoning elided ...]

color 2, connected to 3 (3) and 5 (1). Correct.

So the final answer would be:

{
  "0": 1,
  "1": 2,
  "2": 1,
  "3": 3,
  "4": 2,
  "5": 1,
  "6": 2,
  "7": 3,
  "8": 4,
  "9": 4
}

But let me check if there's another possible coloring. For example, maybe some colors 
could be swapped. But the question just requires any valid coloring. So this should be 
acceptable.
</think>

<answer>
{"0": 1, "1": 2, "2": 1, "3": 3, "4": 2, "5": 1, "6": 2, "7": 3, "8": 4, "9": 4}
</answer>
\end{Verbatim}
\end{tcolorbox}

\begin{tcolorbox}[
    colback=gray!5, colframe=black!60, boxrule=0.5pt,
    fonttitle=\bfseries, title={70\%-Masked variant response},
    breakable,
    width=\linewidth,
    boxsep=1mm, left=1mm, right=1mm, top=1mm, bottom=1mm
]
\begin{Verbatim}[fontsize=\scriptsize]
 3, 3, 2, 4? Wait, vertex 3 is color 3, vertex 2 is color 1, vertex 4 is color 2, vertex
 5 is color 1, vertex 6 is color 1, vertex 7 is color 3, vertex 8 is color 4, vertex 9 
is color 4. Wait, vertex 9 is connected to vertex 3 (color 3), vertex 5 (color 1), 
vertex 0 (color 1), vertex 1 (color 1). So vertex 9 is color 4, which is different from 
all its neighbors. So that's okay.

Wait, but vertex 9 is connected to vertex 3 (color 3), which is okay. Vertex 9 is color 
4. So that's okay.

Wait, but vertex 9 is connected to vertex 1 (color 1), vertex 0 (color 1), vertex 3 
(color 3), vertex 5 (color 1). So vertex 9's color is 4, which is different. So that's 
okay.

So, the color assignments so far

[... 2806 characters of reasoning elided ...]

But this seems to work. Let me just write the JSON as per the example.

The vertices are 0-9. So the JSON would be:

{"0": 1, "1": 2, "2": 1, "3": 3, "4": 2, "5": 1, "6": 2, "7": 3, "8": 4, "9": 4}

But let me check again for vertex 6. Vertex 6 is color 2. It's connected to 3 (color 3) 
and 5 (color 1). So okay.

Yes. So this should be correct.
</think>

<answer>
{"0": 1, "1": 2, "2": 1, "3": 3, "4": 2, "5": 1, "6": 2, "7": 3, "8": 4, "9": 4}
</answer>
\end{Verbatim}
\end{tcolorbox}

\begin{tcolorbox}[
    colback=gray!5, colframe=black!60, boxrule=0.5pt,
    fonttitle=\bfseries, title={Fully Masked variant response},
    breakable,
    width=\linewidth,
    boxsep=1mm, left=1mm, right=1mm, top=1mm, bottom=1mm
]
\begin{Verbatim}[fontsize=\scriptsize]
{"0": 1, "1": 2, "2": 1, "3": 3, "4": 2, "5": 1, "6": 2, "7": 3, "8": 4, "9": 4}
\end{Verbatim}
\end{tcolorbox}


\FloatBarrier

\section{Qualitative analysis of masked-self distillation variants' traces}
\label{app:trace-analysis}
\begin{tcolorbox}[
    colback=gray!5, colframe=black!60, boxrule=0.5pt,
    fonttitle=\bfseries, title={Prompt},
    breakable,
    width=\linewidth,
    boxsep=1mm, left=1mm, right=1mm, top=1mm, bottom=1mm
]

\textit{System:}
\begin{Verbatim}[fontsize=\scriptsize]
You are checking whether a model's generated reasoning trace reads as a coherent,
self-contained chain of thought, or whether it reads as if it is continuing a thought 
that was never stated e.g. it opens using a connective ("So", "Then", "Thus", 
"Given that", "Therefore"), a pronoun or definite reference with no antecedent ("it", 
"this value", "the result", "that number").

You are judging ONLY internal coherence of the trace as a standalone piece of text.
Do not judge correctness of the reasoning or the final answer.
\end{Verbatim}
\vspace{0.3em}
\textit{User:}
\begin{Verbatim}[fontsize=\scriptsize]
Question given to the model: {question}

Model's generated trace (this is the ENTIRE generation, starting from token 1):
{reasoning_trace}

Answer these:   
1. abrupt_start (true/false): does the trace's opening sentence read as though
   picking up mid-reasoning, rather than beginning to address the question?
2. dangling_references (true/false): does the opening 1-2 sentences
   contain a pronoun or definite-reference term (e.g. "it", "this",
   "that value", "the remaining number") with no antecedent stated
    earlier in the trace?
3. coherence_score (1-5): 1 = reads as a truncated fragment / picks up
   mid-thought, 5 = fully self-contained, well-formed opening.

Output JSON only:
{"abrupt_start": true/false, "dangling_references": true/false,
"coherence_score": 1-5}
\end{Verbatim}
\end{tcolorbox}
\captionof{figure}{Prompt used for the LLM-judge coherence audit. The judge
is shown only the question and the trace}
\vspace{0.5em}
\label{fig:coherence-prompt}

We perform a qualitative analysis of traces generated by the Qwen3-8B model and its $\alpha$-masked variants on the $\mathrm{GC}$-$\mathrm{ER}_{\chi_{3-4}}$ test set. To check whether the student-generated traces start abruptly, as if continuing a partially done calculation, we perform an LLM-as-a-judge analysis where we provide the question and the reasoning trace to a coherence audit judge. The judge examines the trace and returns the following: (1) Abrupt start: whether the opening sentence reads as a continuation of an unstated thought rather than a self-contained start. (2) Coherence score: Measuring the coherence of the entire trace. (3) Dangling reference: whether in the opening sentences of the trace, there are reference terms that have no prior antecedent within the trace. We report the coherence audit judge prompt in Figure \ref{fig:coherence-prompt}. 

We use GPT-5-Mini in low reasoning mode as our coherence audit judge. We report our results in Table \ref{tab:coherence-audit}. We see that the audit judge correctly scores the teacher-generated responses and 0\% masked model responses as coherent, with no abrupt starts and dangling references. However, the judge consistently flags the traces generated by the $\alpha$ masked traces as abrupt, incoherent, and having missing references. These preliminary qualitative results suggest that the student models internalize the computations present in the $\alpha$ fraction of the teacher's trace and learn to generate a shorter trace.

\begin{table}[!ht]
  \centering
  \small
  \begin{tabular}{lccccc}
  \toprule
  Metric & Base (think) & 0\% masked & 30\% masked & 50\% masked & 70\% masked \\
  \midrule
  Abrupt-start rate           & 0.00 & 0.00 & 0.94 & 1.00 & 1.00 \\
  Mean coherence score (1--5) & 5.00 & 5.00 & 2.09 & 1.42 & 1.29 \\
  Dangling-reference rate     & 0.00 & 0.00 & 0.24 & 0.91 & 0.90 \\
  \bottomrule
  \end{tabular}
  \caption{LLM-judge coherence analysis of traces generated by the Qwen3-8B model (teacher) and its $\alpha$ masked variants trained on the graph coloring ID test set}
  \label{tab:coherence-audit}
  \end{table}
\end{document}